\documentclass{article}
\usepackage{iclr2026_conference,times}
\usepackage{wrapfig}
\usepackage{enumitem}
\usepackage{amsthm}
\usepackage{graphicx}       
\usepackage{wrapfig}        
\usepackage{subcaption}     
\usepackage{multirow}
\usepackage{booktabs}
\usepackage{adjustbox}
\usepackage{comment}
\usepackage{hyperref}
\usepackage{url}
\usepackage{adjustbox}
\usepackage{float}
\usepackage{booktabs}
 
\usepackage{xcolor}
\newcommand{\gray}[1]{\textcolor{gray}{#1}}

\usepackage{amsmath, amssymb}
\usepackage{amsmath,amsfonts,bm}

\def\eqref#1{equation~\ref{#1}}
\def\1{\bm{1}}

\DeclareMathAlphabet{\mathsfit}{\encodingdefault}{\sfdefault}{m}{sl}
\SetMathAlphabet{\mathsfit}{bold}{\encodingdefault}{\sfdefault}{bx}{n}

\newtheorem{Definition}{Definition}

\title{Dataset Color Quantization: A Training-Oriented Framework for Dataset-Level Compression}

\iclrfinalcopy   

\author{
\textbf{Chenyue Yu}\textsuperscript{3},
\textbf{Lingao Xiao}\textsuperscript{1,2,3},
\textbf{Jinhong Deng}\textsuperscript{1,2,4},
\textbf{Ivor W.\ Tsang}\textsuperscript{1,2,5},
\textbf{Yang He}\textsuperscript{1,2,3}
\\[2mm]
\textsuperscript{1}CFAR, Agency for Science, Technology and Research, Singapore\\
\textsuperscript{2}IHPC, Agency for Science, Technology and Research, Singapore\\
\textsuperscript{3}National University of Singapore \\
\textsuperscript{4}University of Electronic Science and Technology of China (UESTC)\\
\textsuperscript{5}Nanyang Technological University (NTU), Singapore \\[1mm]
\texttt{\{e1143627, xiao\_lingao\}@u.nus.edu, jhdengvision@gmail.com}\\
\texttt{\{Ivor.Tsang, He\_Yang\}@a-star.edu.sg}
}

\begin{document}

\maketitle

\begin{abstract}
Large-scale image datasets are fundamental to deep learning, but their high storage demands pose challenges for deployment in resource-constrained environments. While existing approaches reduce dataset size by discarding samples, they often ignore the significant redundancy within each image -- particularly in the color space. To address this, we propose Dataset Color Quantization (DCQ), a unified framework that compresses visual datasets by reducing color-space redundancy while preserving information crucial for model training. DCQ achieves this by enforcing consistent palette representations across similar images, selectively retaining semantically important colors guided by model perception, and maintaining structural details necessary for effective feature learning. Extensive experiments across CIFAR-10, CIFAR-100, Tiny-ImageNet, and ImageNet-1K show that DCQ significantly improves training performance under aggressive compression, offering a scalable and robust solution for dataset-level storage reduction. 
% Code is available at \href{https://github.com/he-y/Dataset-Color-Quantization}{https://github.com/he-y/Dataset-Color-Quantization}.

\end{abstract}

\section{Introduction}
\label{sec:intro}

The rapid growth of datasets has been pivotal to the success of deep neural networks (DNNs) across diverse applications~\citep{deng2009imagenet}. However, storing and training on such datasets requires significant storage space and computing resources, which poses challenges for deployment on resource-constrained devices such as edge servers, drones, and industrial platforms~\citep{mao2017mobile}. To alleviate this burden, dataset compression methods such as 
dataset pruning~\citep{yang2022dataset} and dataset distillation~\citep{wang2018dataset} have been widely studied. These approaches effectively reduce the number of training samples while maintaining recognition performance, yet they do not explicitly address the \textbf{per-sample storage overhead} that arises from representing full-color images.

In practice, image storage and transmission costs are often dominated by \textbf{color-space redundancy}. For example, images collected from remote monitoring or drone surveillance typically undergo color quantization to reduce bandwidth consumption. While color quantization (CQ; \cite{heckbert1982color}) has been studied extensively for image compression and visualization~\citep{puzicha2000spatial}, existing methods fall short in dataset-level training scenarios. Specifically, CQ can be categorized into two types: 1) \textbf{Image-Property-based CQ}: Preserving human visual perception according to the intrinsic properties of images by mapping the original color space to a reduced set of representative colors~\citep{ozturk2014color}; and 2) \textbf{Model-Perception-based CQ}: Retaining recognizable features to enhance recognition accuracy by pre-trained neural networks~\citep{hou2024scalable}.

\begin{figure*}[t]
    \centering
    \begin{subfigure}[b]{0.45\textwidth}
        \centering
        \includegraphics[width=\linewidth]{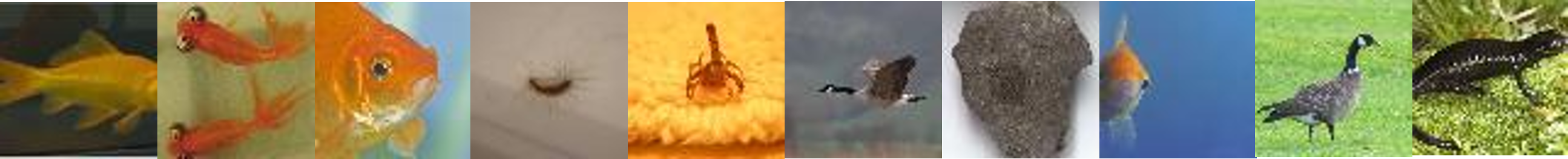}
        \caption{Original Image}
        \label{fig:image1_tinycolor}
    \end{subfigure}
    \hfill
    \begin{subfigure}[b]{0.45\textwidth}
        \centering
        \includegraphics[width=\linewidth]{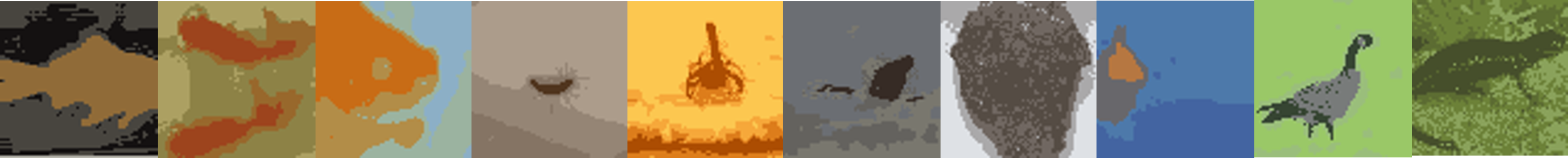}
        \caption{Image-Property-based CQ (K-Means)}
        \label{fig:image2_tinycolor}
    \end{subfigure}

    \vspace{0.5em}  

    \begin{subfigure}[b]{0.45\textwidth}
        \centering
        \includegraphics[width=\linewidth]{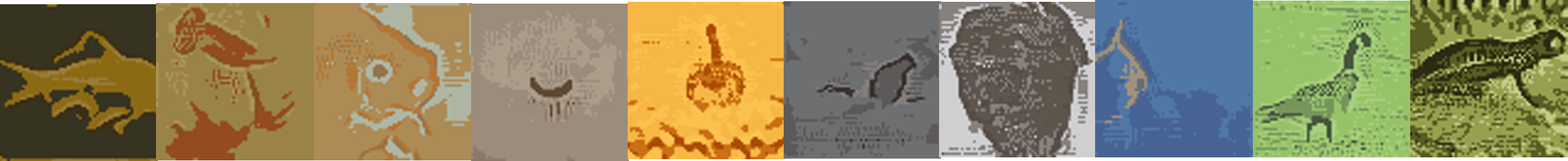}
        \caption{Model-Perception-based CQ (ColorCNN)}
        \label{fig:image3_tinycolor}
    \end{subfigure}
    \hfill
    \begin{subfigure}[b]{0.45\textwidth}
        \centering
        \includegraphics[width=\linewidth]{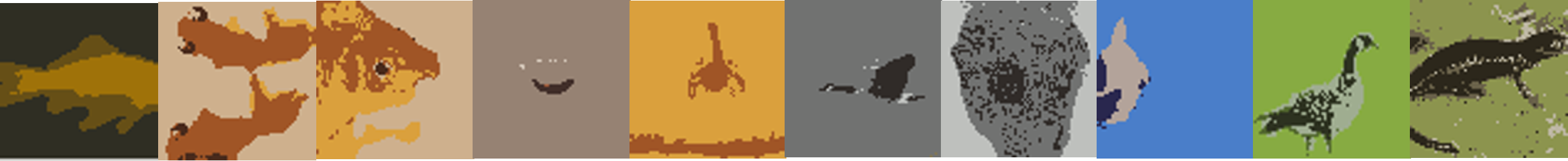}
        \caption{Ours (DCQ)}
        \label{fig:image4_tinycolor}
    \end{subfigure}

    \caption{Visualization of different Color Quantization algorithms on Tiny-ImageNet.
    Images are quantized into 4 colors, which are 2 color bits.
    (a) Original images. 
    (b) Color quantization is performed through K-Means clustering to obtain representative color palettes for each image, wasting bits on backgrounds.
    (c) Independent and representative color palettes obtained by ColorCNN, which have abrupt textural discontinuities.
    (d) Our DCQ assigns more colors to foregrounds and has less textural discontinuity.}
    \label{fig:visualization_tiny}
\end{figure*}

Image-property-based CQ lacks neural network guidance, leading to two issues: (1) ambiguous semantic boundaries and insufficient color contrast~\citep{kimchi2008figure} that hinder discriminative feature learning; (2) uniform bit allocation across all colors, which wastes capacity on background while underrepresenting critical foreground features (Figure~\ref{fig:image2_tinycolor}).  

In contrast, Model-perception-based CQ employs proxy networks to maintain recognition accuracy, but often introduces abrupt texture and edge discontinuities that distort visual features and degrade training performance~\citep{geirhos2018imagenet}. For instance, ColorCNN~\citep{hou2020learning} quantizes CIFAR-10 images into 4 colors (2 bits) and achieves 77\% accuracy under a pre-trained ResNet-18, yet training on the quantized dataset yields only 58\% accuracy due to distorted textures (Figure~\ref{fig:image3_tinycolor}).

To address these limitations, we propose a unified Dataset Color Quantization (DCQ) framework that compresses datasets by reducing redundant color information while preserving both semantic and structural fidelity. Unlike existing methods that quantize images independently or optimize only for inference performance, our approach jointly considers dataset-level consistency, model-aware color significance, and visual structure preservation. Specifically, DCQ clusters images with similar color distributions to enable shared palette learning, prioritizes important regions based on model perception to retain critical semantics, and maintains texture continuity to avoid loss of structural information. This unified strategy achieves compact, quantization-aware datasets that remain effective for downstream training.

Our main contributions are summarized as follows: 1) To our knowledge, this is the first work to propose a solution using a limited set of color palettes to represent datasets, aiming to reduce storage requirements and enable training on color-restricted devices. 2) We introduce a dataset-level color quantization algorithm that combines cluster-shared color palettes, attention-guided bits allocation, and edge-preserving optimization. 3) Extensive experiments on various datasets, including CIFAR-10, CIFAR-100, Tiny-Imagenet, and ImageNet-1K, have validated the effectiveness of our method.  

\section{Related Works}
\label{sec:formatting}
\subsection{Color Quantization}
Color quantization reduces the number of colors used while preserving visual fidelity in RGB space. For maintaining human visualization, cluster algorithms such as MedianCut~\citep{heckbert1982color}, OCTree~\citep{gervautz1988simple}, and K-Means~\citep{cheng2019color} are common cluster algorithms used to group similar colors and represent each group with a single color to find the most representative color palettes; For maintaining neural network's recognition accuracy, ColorCNN~\citep{hou2020learning} utilizes autoencoder to identify significant colors through end-to-end training processing for each image. ColorCNN+~\citep{hou2024scalable} enhances traditional color quantization by combining deep learning and clustering techniques. CQFormer~\citep{su2023name} performs color quantization by mapping input images to quantized color indices through its Annotation Branch, while its Palette Branch identifies key colors in the RGB space. Although these algorithms work well with neural networks and preserve human visual recognition, they do not effectively optimize the train-set for better model performance. 
\subsection{Dataset Pruning}
Dataset Pruning, also known as Coreset Selection, aims to shrink the storage of a dataset by selecting important samples according to some predefined criteria. GraNd/EL2N~\citep{paul2021deep}  quantizes the importance of a sample with its gradient magnitude. TDDS~\citep{zhang2024spanning} uses a dual-depth strategy to achieve a balance between incorporating extensive training dynamics and identifying representative samples for dataset pruning. CCS~\citep{zheng2022coverage} proposed a stratified sampling strategy to increase the classification accuracy in a high pruning ratio. Entropy~\citep{coleman2019selection} proposes selecting the highest-entropy examples for the coreset, as entropy captures the uncertainty of training examples. Forgetting~\citep{toneva2018empirical} refers to the number of instances where an example is misclassified after it has previously been correctly classified during model training and AUM~\citep{pleiss2020identifying} is a data difficulty metric that identifies mislabeled data. However, these algorithms experience a significant accuracy drop at high compression rates, necessitating alternative methods for effective dataset compression under such conditions which means we need to pioneer a new direction to solve this problem, not just reduce the storage by removing images.

\section{Methodology}

\subsection{Motivation}

Dataset pruning and distillation reduce sample counts to lower storage and computation costs, but often overlook per-sample redundancy. In images, storage and transmission are dominated by color redundancy: many pixels share nearly identical colors, especially in smooth regions (e.g., skies, walls) or gradual textures. For instance, a 256×256 RGB image has over 65k pixels but only a few thousand distinct colors. Thus, much capacity is wasted on subtle, semantically irrelevant variations, motivating color quantization or palette-based compression to cut per-image storage while preserving semantic content.

To reduce the color-space redundancy, Color quantization (CQ) has been extensively employed in the contexts of image compression and visualization. However, existing CQ techniques are ill-suited for dataset-level training. Formally, we define color quantization as follows:

\begin{Definition}[Color Space Distribution]
Let $\mathcal{S} \subseteq \mathbb{R}^3$ be the RGB color space. We define two fundamental color distribution representations:

(i) Original Color Palette $\mathcal{P} = \{p_i \in \mathcal{S}\}_{i=1}^m$, where each $p_i$ represents a unique RGB color vector in the continuous color space.
(ii) Quantized Color Palette  $\mathcal{C} = \{c_i \in \mathbb{R}^3\}_{i=1}^k$, where $k \ll m$, and each $c_i$ represents a representative color vector that preserves the most discriminative color features from the original distribution $\mathcal{P}$. Here $d$ denotes the dimensionality of the target color space. The mapping from $\mathcal{P}$ to $\mathcal{C}$ defines a color quantization function:
\[ Q: \mathcal{P} \rightarrow \mathcal{C} \]
where $Q$ performs both cardinality reduction and potential space transformation while maintaining essential color characteristics.
Given an image $I$ with original palette $\mathcal{P}$, the quantized image $I_Q$ is obtained through:
\[ I_Q = Q(I; \mathcal{C}): I \times \mathcal{C} \rightarrow I_Q \]
where $I_Q$ represents the reconstructed image using only colors from palette $\mathcal{C}$.
\label{Def:Color_Distribution}
\end{Definition}

Existing CQ methods can be broadly categorized into two paradigms: 1) Image-Property-Based CQ: This approach reduces color-space redundancy by grouping chromatically similar pixels. While computationally straightforward, it is inherently limited in semantic preservation. Gradual color transitions spanning multiple objects or regions may cause semantically distinct areas to be erroneously merged, resulting in a loss of structural information and critical boundary details. 2) Model-Perception-Based CQ: These methods leverage pre-trained neural networks to guide color allocation~\citep{hou2020learning}, ensuring high recognition accuracy for quantized images. However, they often introduce abrupt texture and edge discontinuities, which can distort visual structure and impair the learning of complete feature representations~\citep{geirhos2018imagenet}. 

Crucially, both paradigms emphasize \textbf{inference-oriented performance}, optimizing the recognition of pre-trained models. Formally, given a dataset $\mathcal{D}=\{(x_i,y_i)\}_{i=1}^N$, a quantization function $Q(\cdot)$, and a fixed pretrained model $f_{\phi}$, inference-oriented quantization optimizes:
\begin{equation}
Q^{\star}
=\arg\max_{Q}\;
\mathbb{E}_{(x,y)\sim\mathcal{D}}
\big[\mathcal{A}(f_{\phi}(Q(x)),y)\big],
\end{equation}
where $\mathcal{A}$ denotes the evaluation metric (\textit{e.g.}, accuracy).

In contrast, our dataset-level color quantization is to optimize the training of high-performance models on quantized datasets. Formally, let $\mathcal{D}_{\text{train}}$ and $\mathcal{D}_{\text{test}}$ denote the training and test sets. Quantization induces a training set $\mathcal{D}_Q=\{(Q(x),y)\}$.
A model $f_{\psi}$ trained on $\mathcal{D}_Q$ is obtained as:
\begin{equation}
   \psi^{\star}(Q)=\arg\min_{\psi}\;
    \mathbb{E}_{(x,y)\sim\mathcal{D}_{\text{Q}}}
    \big[\ell(f_{\psi}(Q(x)),y)\big].
\end{equation}

The training-oriented quantizer is then:
\begin{equation}
    Q^\star_\text{train}
    =\arg\max_{Q}\;
    \mathbb{E}_{(x,y)\sim\mathcal{D}_{\text{test}}}
    \big[\mathcal{A}(f_{\psi^{\star}(Q)}(x),y)\big].
\end{equation}
  \begin{wrapfigure}[20]{r}{0.5\textwidth}   
    \centering

    \begin{subfigure}[b]{\linewidth}
        \centering
        \includegraphics[width=\linewidth]{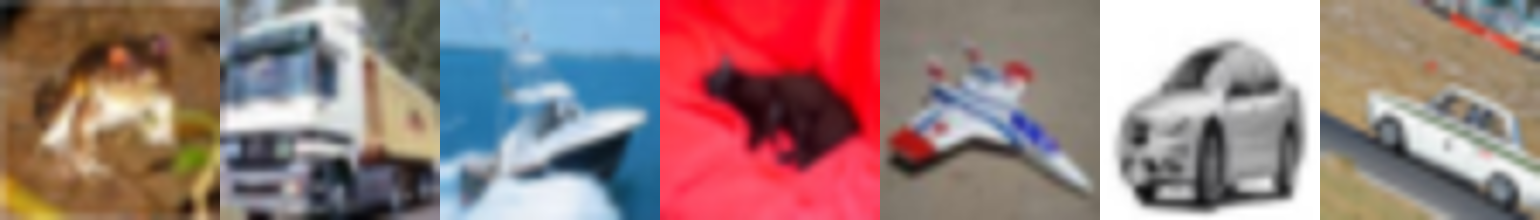}
        \caption{Original Image}
        \label{fig:feature_map_original}
    \end{subfigure}

    \begin{subfigure}[b]{\linewidth}
        \centering
        \includegraphics[width=\linewidth]{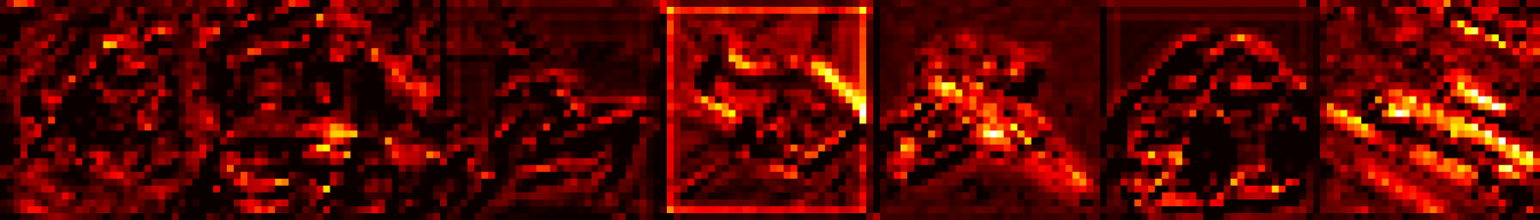}
        \caption{Feature Map from \textbf{First} Layer}
        \label{fig:feature_map_shallow}
    \end{subfigure}

    \begin{subfigure}[b]{\linewidth}
        \centering
        \includegraphics[width=\linewidth]{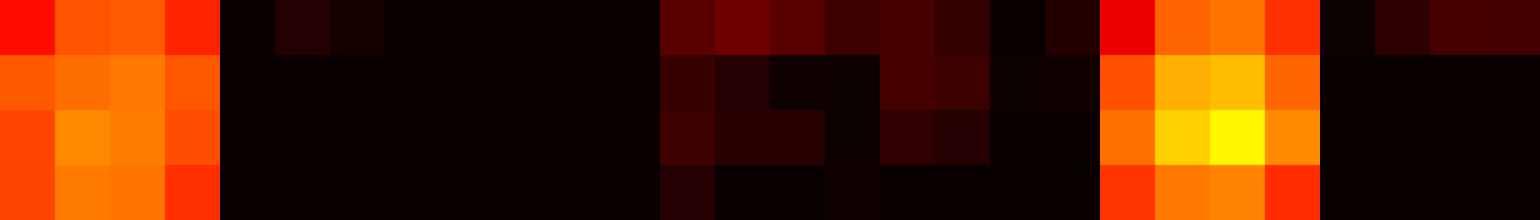}
        \caption{Feature Map from \textbf{Final} Layer}
        \label{fig:feature_map_deep}
    \end{subfigure}
    \caption{Comparison of the original image and feature maps extracted from a ResNet-18 trained on CIFAR-10. Thermal color maps visualize activation strength from black to white, reflecting learned feature hierarchy.}
    \label{fig:wrap_featuremap}
\end{wrapfigure}
To overcome these limitations, we propose a novel \textbf{Dataset Color Quantization (DCQ)} framework, which integrates color quantization into dataset-level compression. Unlike conventional CQ methods that operate independently on individual images, DCQ jointly optimizes palette sharing and semantic preservation across the dataset, achieving a substantial reduction in storage while retaining training efficacy. 
Specifically, we first design a Chromaticity-Aware Clustering (CAC) strategy to aggregate images with chromatically similar distributions into clusters and construct a shared cluster-level palette instead of a sample-wise color palette in previous CQ methods. Then, we propose an Attention-Guided Palette Allocation mechanism that leverages model-derived attention to prioritize color representation for semantically critical regions, ensuring the preservation of essential object features. Moreover, we leverage differentiable quantization to refine the color palette to retain edge and texture fidelity, preserving structural cues crucial for downstream training. Through these mechanisms, DCQ provides a principled approach to dataset-level color compression, effectively minimizing redundant information while safeguarding both semantic and structural content indispensable for high-fidelity model training.

\begin{figure}
    \centering
    \includegraphics[width=1\linewidth]{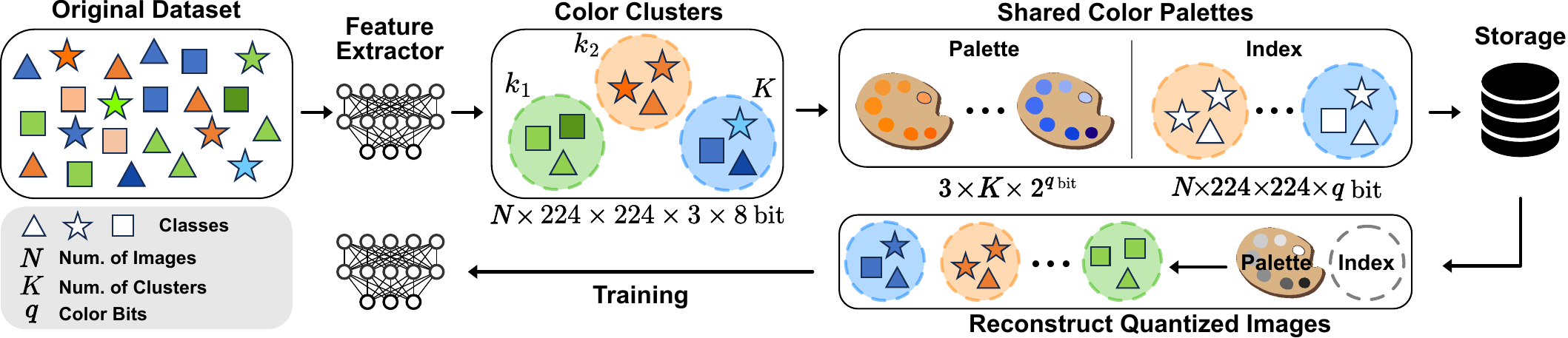}
    \caption{The pipeline of our dataset color quantization framework. First, we apply K-means clustering to group images based on their features extracted from a pre-trained model. Next, within each cluster, we perform K-means on the color palettes of individual images to generate a shared color palette for all images in the same cluster. The generated palettes and their corresponding indices are stored for later use. During training, we retrieve the stored indices and palettes to reconstruct quantized images, which are then used to train a neural network.}
    \label{fig:total-pipline}
\end{figure}

\subsection{Dataset Color Quantization}
The pipeline of the Dataset Color Quantization (DCQ) framework is shown in Figure~\ref{fig:total-pipline}. DCQ first groups images with chromatically similar distributions into clusters and constructs a shared cluster-level palette. Next, within each cluster, we perform K-means on the color palettes of individual images to generate a shared color palette for all images in the same cluster. The generated palettes and their corresponding indices are stored for later use. During training, we retrieve the stored indices and palettes to reconstruct quantized images, which are then used to train a neural network. In the following, we will give a detailed introduction of the proposed method.
\begin{figure}[t]
    \centering

    \begin{subfigure}[b]{0.23\textwidth}
        \centering
        \includegraphics[width=\linewidth]{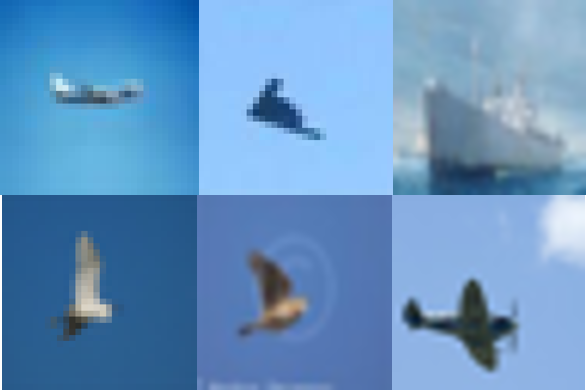}
        \caption{Cluster 1}
        \label{fig:Cluster_0}
    \end{subfigure}
    \hfill
    \begin{subfigure}[b]{0.23\textwidth}
        \centering
        \includegraphics[width=\linewidth]{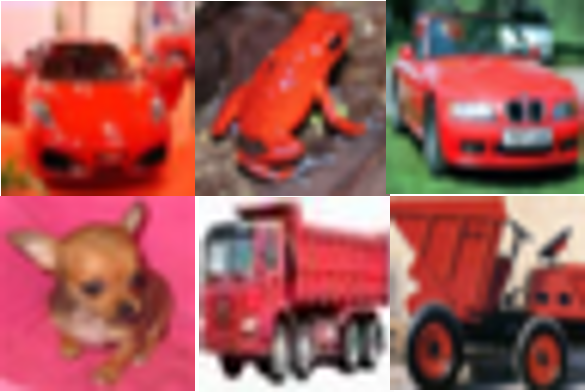}
        \caption{Cluster 2}
        \label{fig:Cluster_1}
    \end{subfigure}
    \hfill
    \begin{subfigure}[b]{0.23\textwidth}
        \centering
        \includegraphics[width=\linewidth]{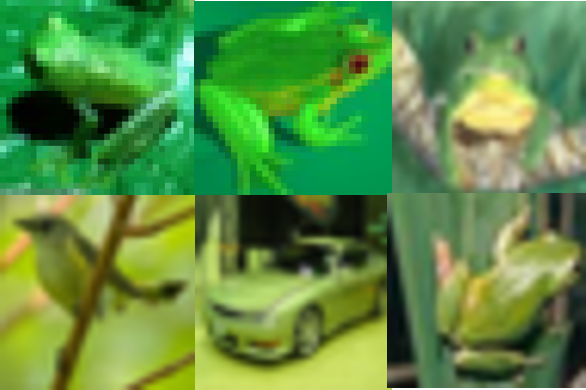}
        \caption{Cluster 3}
        \label{fig:Cluster_2}
    \end{subfigure}
    \hfill
    \begin{subfigure}[b]{0.23\textwidth}
        \centering
        \includegraphics[width=\linewidth]{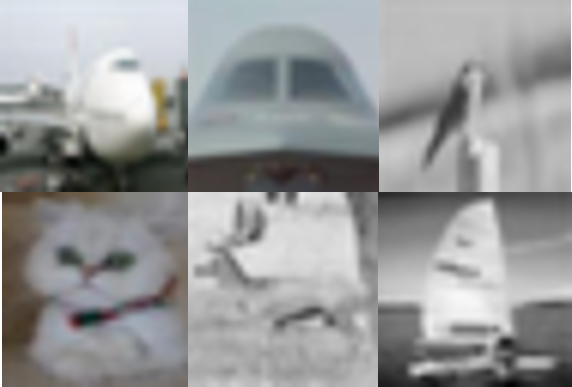}
        \caption{Cluster 4}
        \label{fig:Cluster_3}
    \end{subfigure}

    \caption{Visualization of four clusters obtained by applying ResNet-18 first-block features to partition CIFAR-10 into 20 clusters.}
    \label{fig:wrap_visualization_cluster}
\end{figure}
\vspace{1.8em}

\subsubsection{Chromaticity-Aware Clustering}
\label{sec:deep-cluster}
Traditional CQ methods typically construct palettes independently for each image. This per-image quantization strategy allows the color palette to be optimized to the specific distribution of a single image, thereby minimizing quantization error within that image. However, such independence introduces inconsistency across the dataset: visually similar regions in different images may be mapped to disparate representative colors. As a result, the color space perceived by the training model becomes fragmented, which increases the difficulty of learning stable semantic boundaries and weakens the model’s generalization ability.

To overcome this limitation, we propose the use of shared palettes across multiple images. By quantizing a collection of images rather than on individual samples, the learned palettes achieve greater cross-image consistency. This consistency reduces the semantic ambiguity induced by palette misalignment and enables the model to perceive color distributions more stably. Nevertheless, constructing a single palette from a large set of diverse images poses a new challenge. When the overall color distribution becomes too broad, a fixed-size palette must accommodate a wider variety of colors. Consequently, the quantization error for individual images increases, as the palette can no longer precisely capture each image’s local chromatic distribution.

To this end, we introduce Chromaticity-Aware Clustering (CAC) mechanism that first partitions the dataset into groups of images with similar color distributions to balance the trade-off between cross-image consistency and quantization fidelity. Shared palettes are then generated within each cluster, allowing the method to retain cross-image consistency while constraining the color variance that the palette must represent. In this manner, quantization error remains controlled, and the resulting palettes better preserve semantic boundaries and structural features across images.

In particular, to identify images with similar color distributions, we utilize semantic feature maps  $\psi_i(x)$, which are the outputs of the $i$-th layer in a neural network model $\mathcal{M}$. These feature maps capture hierarchical representations of an image, from low-level color patterns to high-level semantics. The resulting feature vector provides a comprehensive representation of the image’s color distribution, enabling effective comparison across images.
 While deeper output feature maps $\psi_i(x)$ encode abstract semantic concepts like object parts and contextual relationships, shallow layers capture local patterns including color distributions~\citep{zeiler2014visualizing}.

Through empirical analysis of feature distributions across layers, we observe an inherent trade-off:
\begin{equation}\label{eq:tradeoff}
    \text{as } i \uparrow: \begin{cases}
        \text{Sem}(\psi_i) \uparrow & \text{(semantic abstraction)} \\
        \text{Vis}(\psi_i) \downarrow & \text{(visual fidelity)},
    \end{cases}
\end{equation}
where $\text{Sem}(\cdot)$ represents semantic information capacity and $\text{Vis}(\cdot)$ indicates visual fidelity, and the visualization of feature-map is shown in Figure \ref{fig:wrap_featuremap}. Based on this observation, we select the shallow layer feature map as our perceptual representation:
\begin{equation}\label{eq:choice}
    \phi_{\mathcal{M}}(x) = \psi_{shallow}(x)
\end{equation}
 We partition the dataset using K-Means clustering on semantic features $\psi_{shallow}(x)$ to create $k$ clusters. We set $k=20$ through ablation experiments. As shown in Figure \ref{fig:wrap_visualization_cluster}, this clustering not only groups images by dominant color distributions (blue, red, green, and gray tones) but also transcends class boundaries by grouping semantically distinct objects with similar chromatic properties. 
\begin{figure}[ht]
    \centering
    \begin{subfigure}[c]{0.33\textwidth}
        \includegraphics[width=\textwidth]{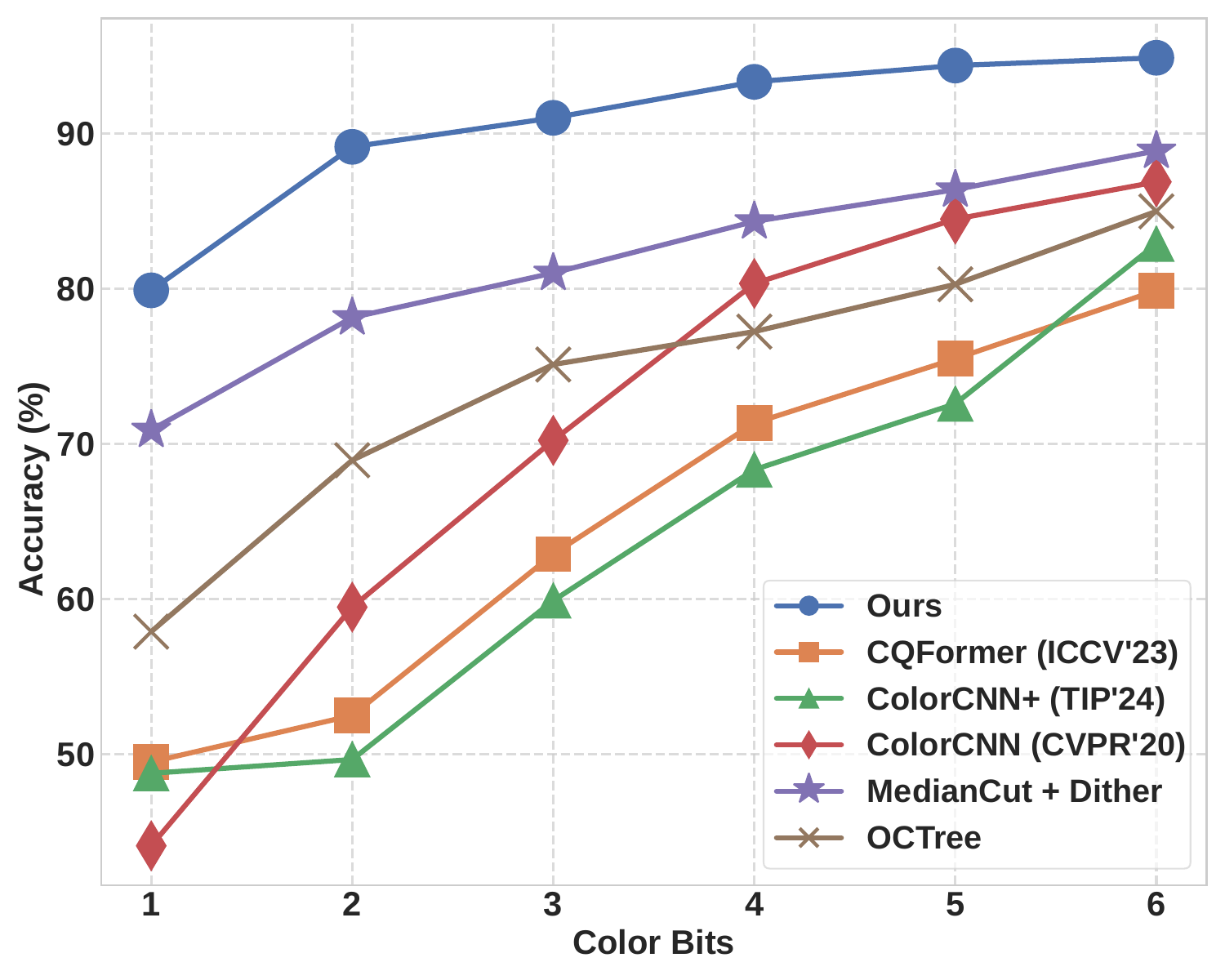}
        \caption{CIFAR-10.}
    \end{subfigure}\hfill
    \begin{subfigure}[c]{0.33\textwidth}
        \includegraphics[width=\textwidth]{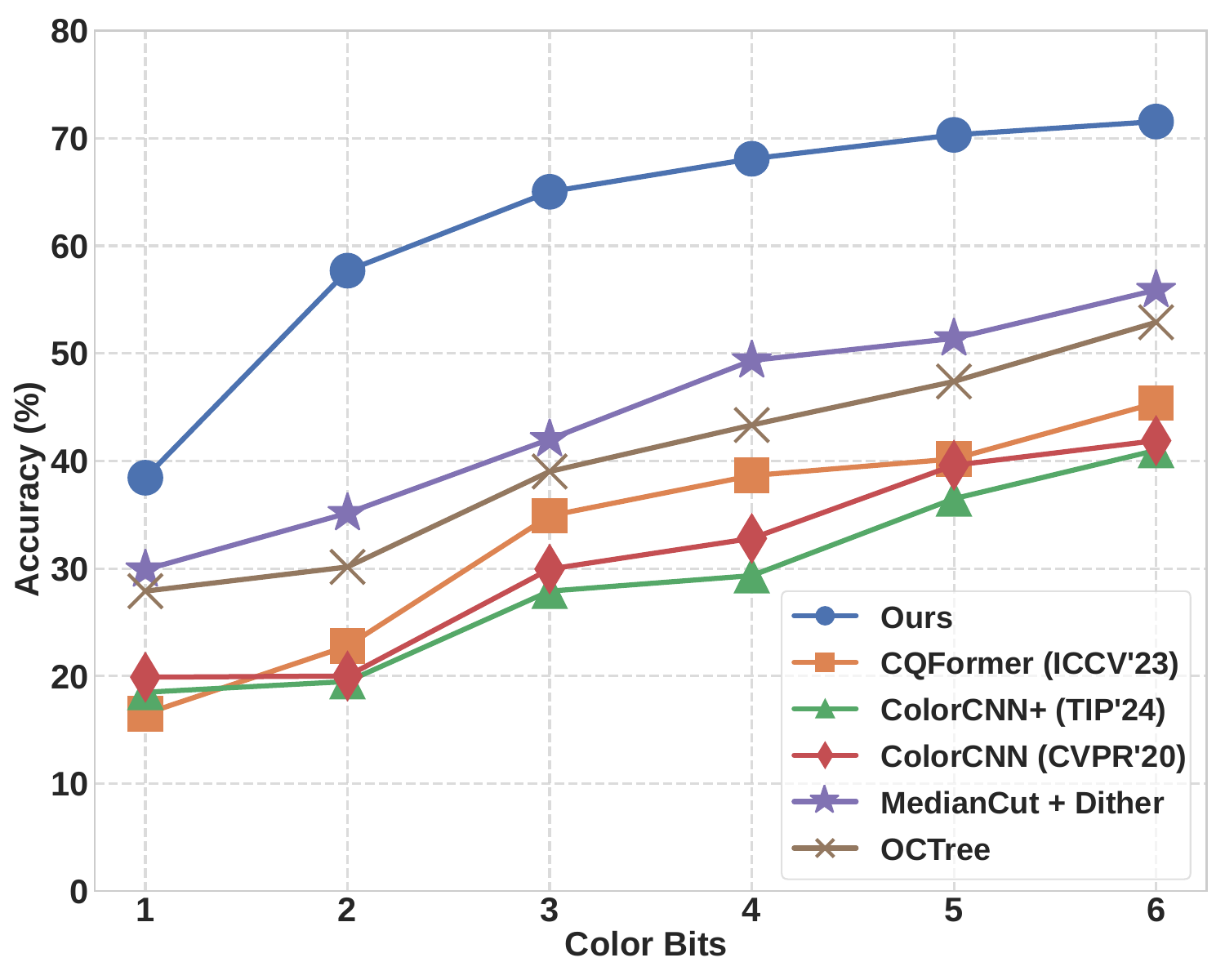}
        \caption{CIFAR-100.}
    \end{subfigure}\hfill
    \begin{subfigure}[c]{0.33\textwidth}
        \includegraphics[width=\textwidth]{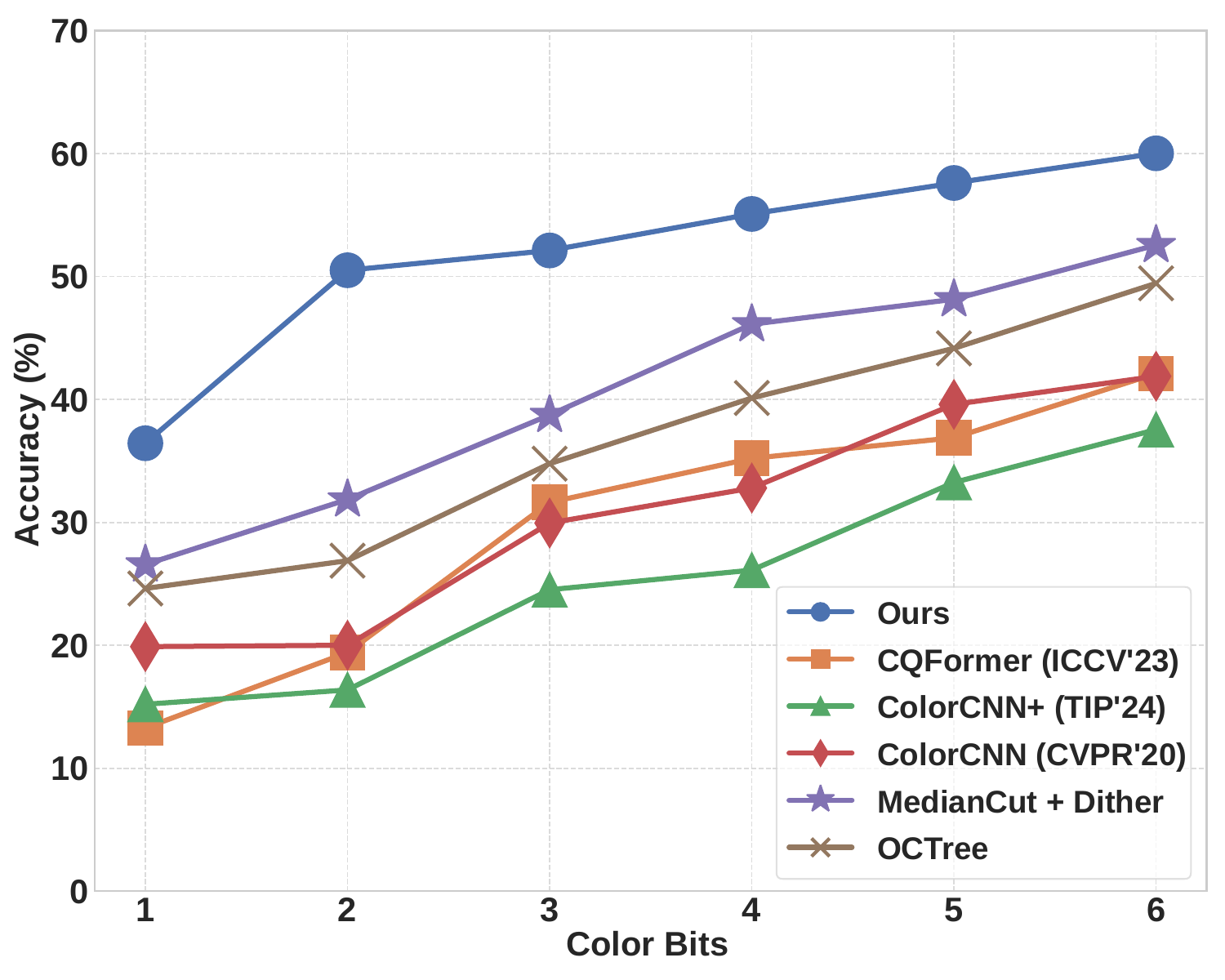}
        \caption{Tiny-ImageNet.}
    \end{subfigure}
    \caption{Comparison between prior color quantization methods and our approach, evaluated with ResNet-18. Detailed results are provided in Appendix~\ref{appendix:primary-cq}. Unlike the original paper, which quantized the test set while training on the original train set, we quantize the entire train-set and keep the test-set unchanged.
    }
    \label{fig:training}
\end{figure}
\subsubsection{Attention-Guided Palette Allocation}
\label{sec:gradcam}

Traditional color quantization approaches typically assume uniform importance across color palettes~\citep{celebi2023forty}. However, different image regions contribute unequally to semantic recognition, and allocating bits uniformly across all regions may neglect critical structures. To address this, we introduce the notion of \textit{Color Palette Impact Level}, which distinguishes palettes that are crucial for model performance from those that are less influential.

\begin{Definition}[Color Palette Impact Level]
We categorize the roles of color palettes in model recognition into two distinct classes: (1) \textbf{High-Impact Palettes} ($P_H$): These palettes significantly influence model performance. Changing them to any other palette results in a noticeable increase in loss. (2) \textbf{Low-Impact Palettes} ($P_L$): These palettes have little effect on model performance. There exists at least one alternative palette that can replace them without a meaningful change in loss.
\label{Def:Color_Palette_Impact_Level}
\end{Definition}

To identify high-impact palettes, we employ Grad-CAM++~\citep{chattopadhay2018grad} to obtain an attention map generated from task-specific pre-trained models. The Grad-CAM++ heatmap reveals discriminative regions that contribute to the network's classification decision. The higher attention values indicate greater relevance to the network’s prediction. For each image, we retain the top $k_{Gra}\%$ of pixels with the highest attention values, ensuring that only the most relevant regions are preserved, while all remaining pixels are set to zero. The value of $k_{Gra}\%$ was determined through ablation studies in Appendix \ref{appendix:Grad_CAM}. 

The selected pixels from all images within a cluster are aggregated to form an expanded palette space. Following prior work~\citep{orchard1991color}, we convert the RGB values of these pixels to LAB color space to better preserve perceptual similarity. Finally, K-means clustering is applied to this aggregated LAB palette space to generate a shared quantized palette that balances compactness and semantic fidelity across images.

\subsubsection{Texture-preserved Palette Optimization.}
\label{sec:texture}
A key goal of color quantization is to preserve important texture information. However, traditional K-Means clustering assigns pixels based solely on color similarity, ignoring the image’s structural details. Without an error feedback mechanism to measure  Texture Loss (TL), particularly in local gradients, edges, and spatial patterns, this approach often leads to texture degradation, especially in fine-detail regions. To address this, our algorithm refines K-Means-generated color palettes through a quantizable optimization process, reducing  Texture Loss (TL) and better preserving local structural details.

Inspired by style transfer~\citep{johnson2016perceptual}, we designed a differentiable textural palette optimization model that performs differentiable color quantization by mapping each pixel in the image to its nearest color in a palette, applying a straight-through estimator (STE)~\citep{yin2019understanding} to enable backpropagation through the quantization process. For each cluster, after obtaining the shared color palette, we optimize the palette using a gradient descent algorithm to minimize the loss function, ensuring that the color differences between the original and quantized images are minimized. We optimize the color palette by minimizing the edge distribution differences $EL$ between the original image and image after color quantization which can be calculated using:
\begin{equation}
\begin{aligned}
    G(I) &= \sqrt{(I \ast S_x)^2 + (I \ast S_y)^2}, \\
    \text{$EL$} &= \sum_{i=1}^3 w_i \cdot \text{MSE} \Big( G(I_{\text{orig}}^i), G(I_{\text{quant}}^i) \Big),
\end{aligned}
\label{eq:total_loss}
\end{equation}
 
where $I_{\text{orig}}^i$ and $I_{\text{quant}}^i$ represent the $i$-th channel (L, A, B) of the original and quantized images, $S_x$ and $S_y$ using Sobel operator~\citep{kanopoulos1988design} to calculate edge information for each channel, MSE stands for the mean squared error and $w_i$ is the weights for each channel and $\ast$ stands for convolution, adjusting the contribution of each channel to the total loss. The result of using Texture-preserved Palette Optimization was shown as ablation experiments in Appendix~\ref{appendix:sobel}.  

Figure \ref{fig:visualization_tiny} shows the result of our dataset color quantization algorithm. Under 2-bit color quantization, compared to traditional Image-Property-based quantization, our approach achieves superior edge-texture preservation and palette allocation efficiency, resulting in enhanced structural integrity and detail fidelity. Compared to Model-Perception-based quantization, our method better preserves perceptual fidelity with enhanced texture continuity.

\section{Experiments}
 
\subsection{Experiment Settings}

In this task, the effectiveness of the proposed dataset color quantization is evaluated from popular datasets: CIFAR-10, CIFAR-100~\citep{krizhevsky2009learning}, Tiny-ImageNet~\citep{le2015tiny} and ImageNet-1K~\citep{deng2009imagenet} with ResNet~\citep{he2016deep}.
Details of datasets are provided in the Appendix~\ref{appendix:dataset}. In our algorithm, we used our color quantization algorithm on the original train-set and kept original test-set. 
Since our work tackles a less-studied problem of dataset color quantization with no known clear solution, it is important to set an adequate baseline for comparison.
We choose two aspects of the most relevant baselines to show the efficiency of our algorithms.
(1) For \textbf{color quantization}, we choose ColorCNN~\citep{hou2020learning}, ColorCNN+~\citep{hou2024scalable}, CQFormer~\citep{su2023name}, MedianCut~\citep{heckbert1982color} and OCTree~\citep{gervautz1988simple}.
We apply image color quantization algorithms to every image in the entire train-set while keeping the test-set unchanged. This differs from the original paper, which quantized the test-set while training on the original train-set.
(2) For \textbf{dataset pruning} algorithms, EL2N~\citep{paul2021deep}, Entropy~\citep{coleman2019selection} and  Forgetting~\citep{toneva2018empirical}, CCS~\citep{zheng2022coverage}, TDDS~\citep{zhang2024spanning} are used as baselines. These algorithms use a coreset from the train-set to train a model and evaluate it on the original test-set.
More information about baselines is provided in the Appendix~\ref{appendix:baseline}.
The compression ratio achieved through color quantization can be formally defined in relation to the bit depth reduction.
Starting from the standard 24-bit RGB color space (8 bits per channel), when quantizing to $q$ bits, the color palette is reduced to $2^q$ distinct colors.
The corresponding compression ratio $q_r$ is given by $q_r = 1 - q/24$, establishing a direct mapping between the compression levels of color quantization and dataset pruning strategies.  
 \begin{table}[h]
    \centering
    \small   
    \setlength{\tabcolsep}{3pt}   
    \caption{Comparison of dataset pruning algorithms and our dataset color quantization algorithm on CIFAR-10, CIFAR-100 with ResNet-18, and ImageNet-1K with ResNet-34.}
    \vspace{0.3em}

    \resizebox{\linewidth}{!}{
    \begin{tabular}{l|ccccc|ccccc|ccccc}
        \toprule
        & \multicolumn{5}{c|}{CIFAR-10} & \multicolumn{5}{c|}{CIFAR-100} & \multicolumn{5}{c}{ImageNet-1K} \\
        Color Bits & 5 & 4 & 3 & 2 & 1 & 5 & 4 & 3 & 2 & 1 & 5 & 4 & 3 & 2 & 1 \\
        Colors/Image & 32 & 16 & 8 & 4 & 2 & 32 & 16 & 8 & 4 & 2 & 32 & 16 & 8 & 4 & 2 \\
        Prune Ratio & 80\% & 83\% & 87.5\% & 92\% & 96\% & 80\% & 83\% & 87.5\% & 92\% & 96\% & 80\% & 83\% & 87.5\% & 92\% & 96\% \\
        \midrule
        Random & 88.15 & 84.38 & 80.15 & 77.04 & 70.08 & 57.36 & 50.96 & 42.19 & 39.71 & 36.68 & 62.54 & 58.18 & 53.34 & 50.32 & 20.28 \\
        Entropy & 79.08 & 74.59 & 67.58 & 64.51 & 57.06 & 41.83 & 35.45 & 29.77 & 28.96 & 22.16 & 55.80 & 44.59 & 42.04 & 31.04 & 17.06 \\
        EL2N & 70.32 & 67.06 & 25.99 & 21.31 & 19.85 & 15.51 & 12.51 & 10.36 & 8.36 & 7.51 & 31.22 & 21.28 & 13.99 & 11.69 & 9.81 \\
        AUM & 57.84 & 49.09 & 29.58 & 25.60 & 21.35 & 16.38 & 11.83 & 9.45 & 8.77 & 5.37 & 21.12 & 15.09 & 10.13 & 8.93 & 6.35 \\
        CCS$_\mathrm{AUM}$ & 90.53 & 88.97 & 87.61 & 75.01 & 73.02 & 63.19 & 60.05 & 52.16 & 30.02 & 28.24 & 64.49 & 58.44 & 57.96 & 45.58 & 31.31 \\
        CCS$_\mathrm{Forg.}$ & 90.93 & 88.05 & 86.66 & 74.31 & 73.02 & 63.99 & 60.45 & 51.86 & 31.12 & 27.33 & 65.01 & 58.74 & 57.57 & 46.28 & 30.44 \\
        CCS$_\mathrm{EL2N}$ & 89.81 & 88.47 & 87.01 & 74.77 & 73.17 & 61.83 & 60.75 & 52.16 & 32.07 & 28.17 & 64.71 & 58.15 & 56.54 & 45.19 & 29.14 \\
        TDDS (Strategy-E) & 91.30 & 88.72 & 87.47 & 77.32 & 72.46 & 63.01 & 61.44 & 55.13 & 32.15 & 26.55 & 62.56 & 56.48 & 54.91 & 43.91 & 29.56 \\
        \textbf{DCQ (Ours)} & \textbf{94.39} & \textbf{93.34} & \textbf{91.02} & \textbf{89.15} & \textbf{79.90} & \textbf{69.05} & \textbf{67.89} & \textbf{65.02} & \textbf{57.69} & \textbf{38.44} & \textbf{66.99} & \textbf{64.34} & \textbf{62.02} & \textbf{49.69} & \textbf{35.95} \\
        \midrule
        Full Accuracy & \multicolumn{5}{c|}{95.45\%} & \multicolumn{5}{c|}{78.21\%} & \multicolumn{5}{c}{73.54\%} \\
        \bottomrule
    \end{tabular}
    }
    \label{tab:pruning_methods_combined}
\end{table}

\begin{table}[t]
    \centering
    \small
    \caption{Ablation Studies of Different Strategies.}
    \begin{adjustbox}{max width=\linewidth}
    \begin{tabular}{ccc}
         
        \begin{subtable}[t]{0.40\textwidth}
            \centering
            \setlength{\tabcolsep}{0.19em}
            \caption{Clustering Features Comparison.}
            \begin{tabular}{cccccc}
                \toprule
                \textbf{Bits} & 
                \textbf{Label} & 
                \textbf{Random} &
                \textbf{Image} & 
                \textbf{Final} &
                \textbf{Shallow} \\
                \midrule
                1 & 40.10 & 28.44 & 68.44& 42.10 & \textbf{79.90} \\
                2 & 51.08 & 39.65 & 79.15& 53.78 & \textbf{89.15} \\
                3 & 65.33 & 53.21 & 81.21& 66.39 & \textbf{91.02} \\
                4 & 74.05 & 61.33 & 84.33& 75.15 & \textbf{93.34} \\
                5 & 79.42 & 70.29 & 89.29& 80.44 & \textbf{94.39} \\
                6 & 81.99 & 75.52 & 90.52& 83.93 & \textbf{94.89} \\
                \bottomrule
            \end{tabular}
            \label{tab:cluster_features}
        \end{subtable}

        &

        \begin{subtable}[t]{0.58\textwidth}
            \centering
            \footnotesize
            \setlength{\tabcolsep}{0.51em}
            \caption{Cluster Numbers on CIFAR-10 (First Level clustering).}
            \begin{tabular}{ccccccc}
                \toprule
                \textbf{Bits} & 1 & 2 & 3 & 4 & 5 & 6 \\
                \midrule
                1 cluster & 71.02 & 84.52 & 86.71 & 89.49 & 90.55 & 91.01\\
                10 cluster & 76.12 & 87.95 & 90.08 & 92.14 & 92.79 & 93.69 \\
                \textbf{20 cluster} & \textbf{79.90} & \textbf{89.15} & \textbf{91.02} & \textbf{93.34} & \textbf{94.39} & \textbf{94.89} \\
                50 cluster & 77.34 & 87.23 & 90.11 & 92.53 & 92.41 & 93.05 \\
                100 cluster & 76.94 & 86.61 & 90.91 & 91.83 & 92.66 & 93.11 \\
                150 cluster & 76.77 &  86.21 & 87.99 &  90.04 & 92.31 & 93.58 \\
                \bottomrule
            \end{tabular}
            \label{tab:num_clusters}
        \end{subtable}

        &

        % --- Subtable C (unchanged) ---
        \begin{subtable}[t]{0.41\textwidth}
            \centering
            \setlength{\tabcolsep}{0.39em}
            \caption{Attention Map Methods (Second level clustering).}
            \begin{tabular}{ccccc}
                \toprule
                \textbf{Bits} & 1 & 2 & 3 & 4 \\
                \midrule
                None & 72.02 & 84.02 & 86.79 & 89.09 \\
                GradCAM & 75.05 & 85.75 & 88.02 & 90.55 \\
                \textbf{GradCAM++} & 79.90  & \textbf{89.15} &  91.02 & \textbf{93.34} \\
                RISE & 79.04 & 88.91 & \textbf{92.05} & 93.32 \\
                LayerCAM & \textbf{80.66} & 89.71 & 90.95 & 93.09 \\
                \bottomrule
            \end{tabular}
            \label{tab:attention_map}
        \end{subtable}
    \end{tabular}
    \end{adjustbox}
    \label{tab:ablation_all}
\end{table} 

\textbf{Comparison to Color Quantization.} Figure \ref{fig:training} shows the performance of using the color quantized dataset for training.
Our performance substantially outperforms other methods from 1 to 6 bits for all three datasets including (a) CIFAR-10, (b) CIFAR-100, and (c) Tiny-ImageNet. 
Using a 2-bit color palette (4 colors) for quantization, our method (DCQ) achieves accuracies of 89.15\% on CIFAR-10, 57.69\% on CIFAR-100, and 50.51\% on Tiny-ImageNet. This represents a improvement over the ColorCNN~\citep{hou2020learning} results, with gains of 30.00\% (89.15\% - 59.15\%) on CIFAR-10, 35.37\% (57.69\% - 22.32\%) on CIFAR-100, and 18.06\% (50.51\% - 32.45\%) on Tiny-ImageNet.

\textbf{Comparison to Dataset Pruning.} Table \ref{tab:pruning_methods_combined} presents the performance comparison of various dataset pruning methods against our dataset color quantization (DCQ) algorithm. DCQ consistently outperforms other methods across all pruning ratios and datasets. For instance, under a 96\% compression ratio, we apply 1-bit quantization (2 colors per image), clustering the entire dataset into 20 
clusters. As a result, only 40 distinct colors are used to 
quantize the entire train-set. DCQ achieves accuracies of 79.9\% on CIFAR-10 and 38.44\% on CIFAR-100, significantly surpassing the CCS results of 73.02\% and 28.24\%, respectively. These results highlight DCQ as an effective color quantization strategy, particularly under high pruning ratios.
Similarly, DCQ excels on ImageNet-1K under high pruning ratios. 
 
\subsection{Ablation Studies} 
\begin{wraptable}[11]{r}{0.5\textwidth}
    \centering
    \scriptsize
    \setlength{\tabcolsep}{0.3em}
    \caption{Comparison of dataset pruning algorithms and our algorithm on CIFAR-10 with ResNet-34 and ResNet-50.
    }
    \resizebox{0.99\linewidth}{!}{
    \begin{tabular}{l|ccccc|ccccc}
        \toprule
        & \multicolumn{5}{c|}{ResNet-34, CIFAR-10} & \multicolumn{5}{c}{ResNet-50, CIFAR-10} \\
        & 80\% & 83\% & 87.5\% & 92\% & 96\% & 80\% & 83\% & 87.5\% & 92\% & 96\% \\
        \midrule
        Random & 86.13 & 83.48 & 76.19 & 74.44 & 65.08 & 87.03 & 83.18 & 79.19 & 75.44 & 68.08 \\
        EL2N & 70.21 & 66.36 & 24.69 & 23.11 & 22.19 & 69.71 & 65.96 & 24.59 & 23.31 & 21.89 \\
        AUM & 58.34 & 49.59 & 29.98 & 21.60 & 20.35 & 57.84 & 49.09 & 29.58 & 25.60 & 21.35 \\
        CCS$_\mathrm{AUM}$ & 89.34 & 87.92 & 87.52 & 74.01 & 71.42 & 89.44 & 88.11 & 86.52 & 74.31 & 71.02 \\
        TDDS & 89.58 & 88.15 & 86.92 & 73.05 & 69.48 & 90.68 & 88.55 & 87.62 & 74.01 & 70.46 \\
        \textbf{Ours} & \textbf{94.39} & \textbf{93.14} & \textbf{91.15} & \textbf{89.47} & \textbf{79.87} & \textbf{94.19} & \textbf{92.94} & \textbf{90.55} & \textbf{88.27} & \textbf{77.26} \\
        \bottomrule
    \end{tabular}
    }
    \label{tab:pruning_methods_cifar10_combined}
\end{wraptable} 
\textbf{How Many Clusters?}
Table \ref{tab:num_clusters} presents the impact of varying the number of clusters in the initial K-Means clustering step.
Using the least number of clusters, which is 1 cluster, extracts all color palettes to get a shared quantized color palette for every image in the dataset.
On the other hand, the most number of clusters (i.e., 50,000 clusters for CIFAR-10) means that we assign each image a unique quantized color palette, which can be seen as Image-Property-based CQ.
The best performance is achieved with 20 clusters.

\textbf{Which Feature to Use?}
Table \ref{tab:cluster_features} compares clustering performance across different feature extraction approaches: (1)  label-based grouping that clusters images within the same class; (2) randomly assign images into different clusters; (3)  direct image clustering that applies K-Means on raw pixel values; (4) final-layer feature maps from the final residual block and (5) shallow-layer feature maps from the first residual block as the different baselines. The shallow-layer features consistently outperform all alternatives, which aligns with our analysis in Sec.~\ref{sec:deep-cluster}. Table \ref{tab:cluster_features} shows that clustering with Shallow-layer Feature maps achieves the best results compared to other features.

\textbf{Which Method to Get Attention Maps?}
Table~\ref{tab:attention_map} compares different attention map generation methods, including Grad-CAM~\citep{selvaraju2016grad}, LayerCAM~\citep{jiang2021layercam}, and RISE~\citep{petsiuk2018rise}. Our algorithm is guided by attention maps and is not limited to Grad-CAM++; other methods are equally applicable.
\begin{figure}[H]
    \centering
    \begin{subfigure}[c]{0.33\textwidth}
        \includegraphics[width=1\linewidth]{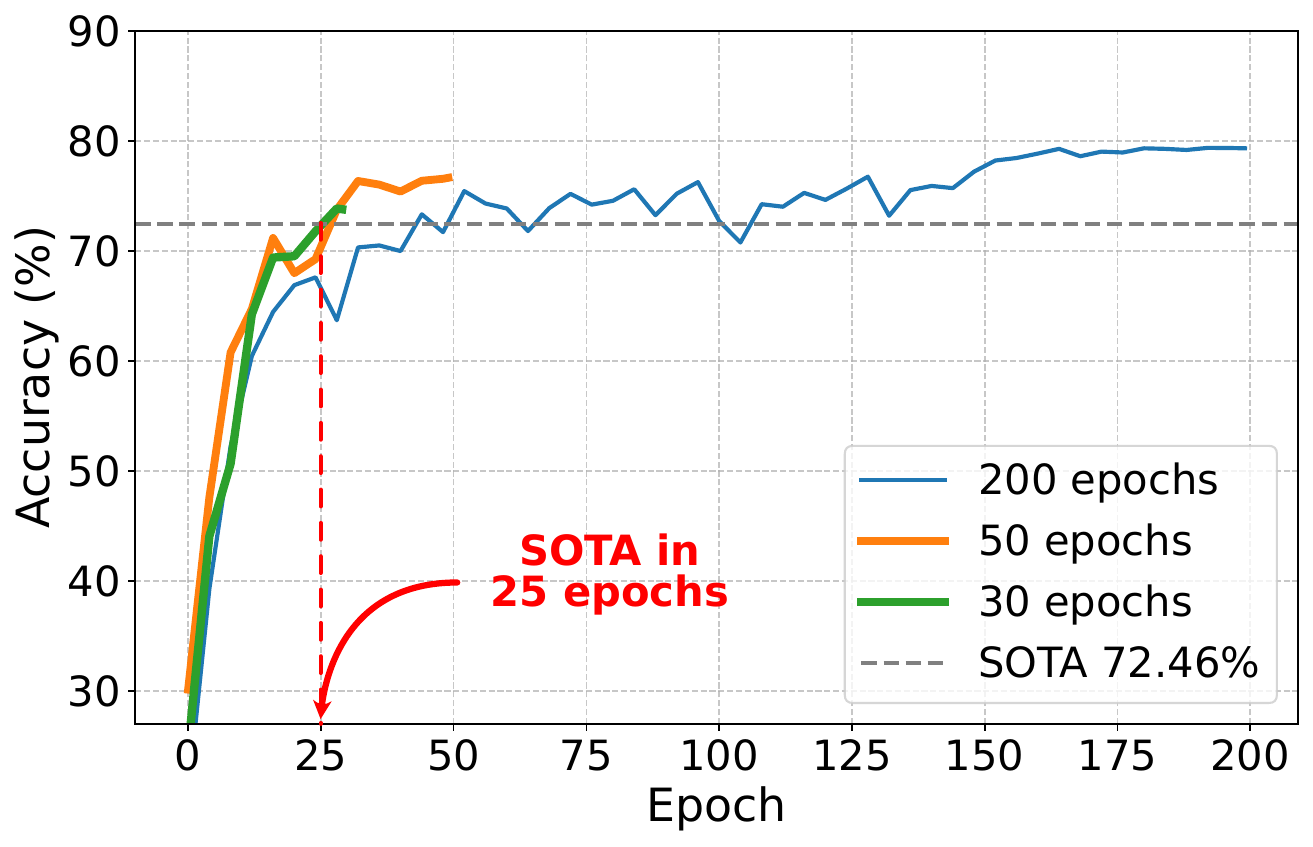}
        \caption{1 Bit (2 colors).}
    \end{subfigure}\hfill
    \begin{subfigure}[c]{0.33\textwidth}
        \includegraphics[width=1\linewidth]{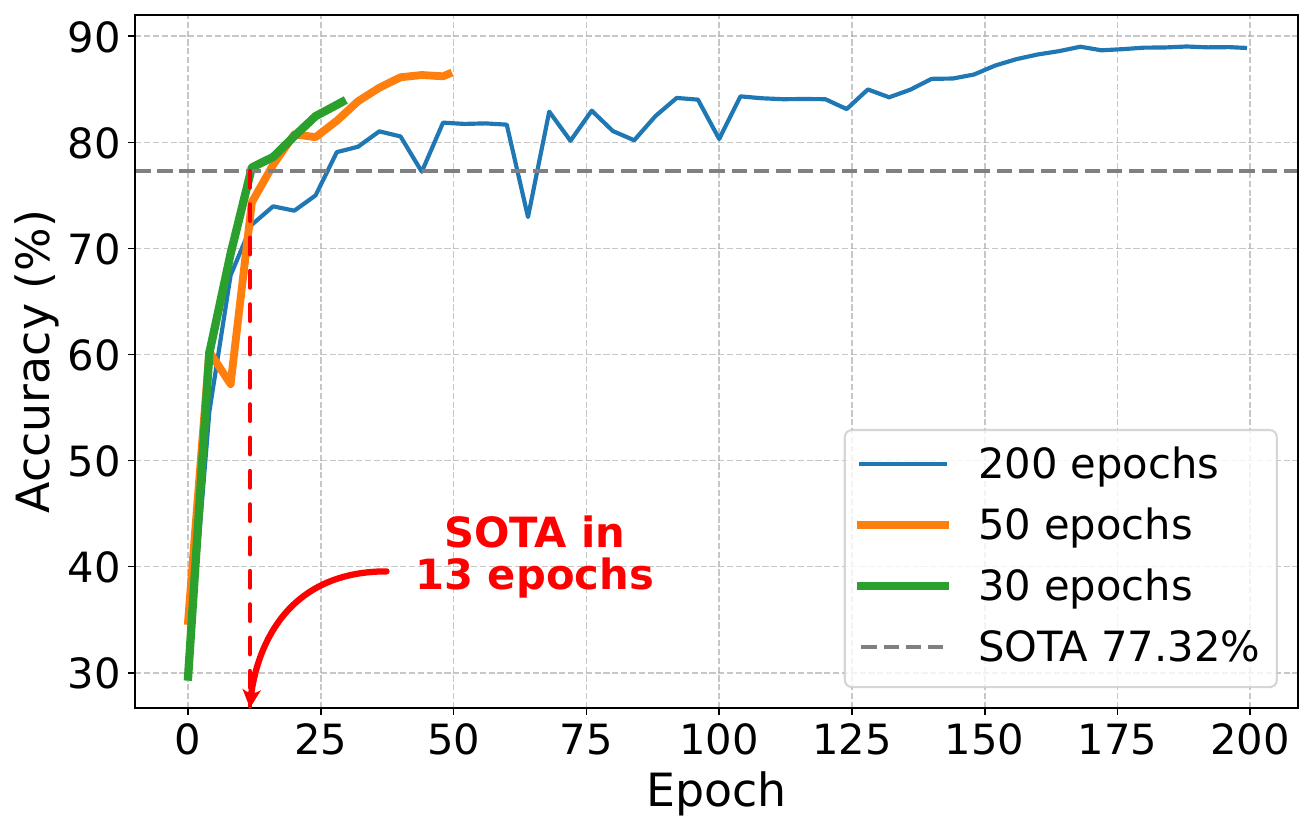}
        \caption{2 Bit (4 colors).}
    \end{subfigure}
    \begin{subfigure}[c]{0.33\textwidth}
        \includegraphics[width=1\linewidth]{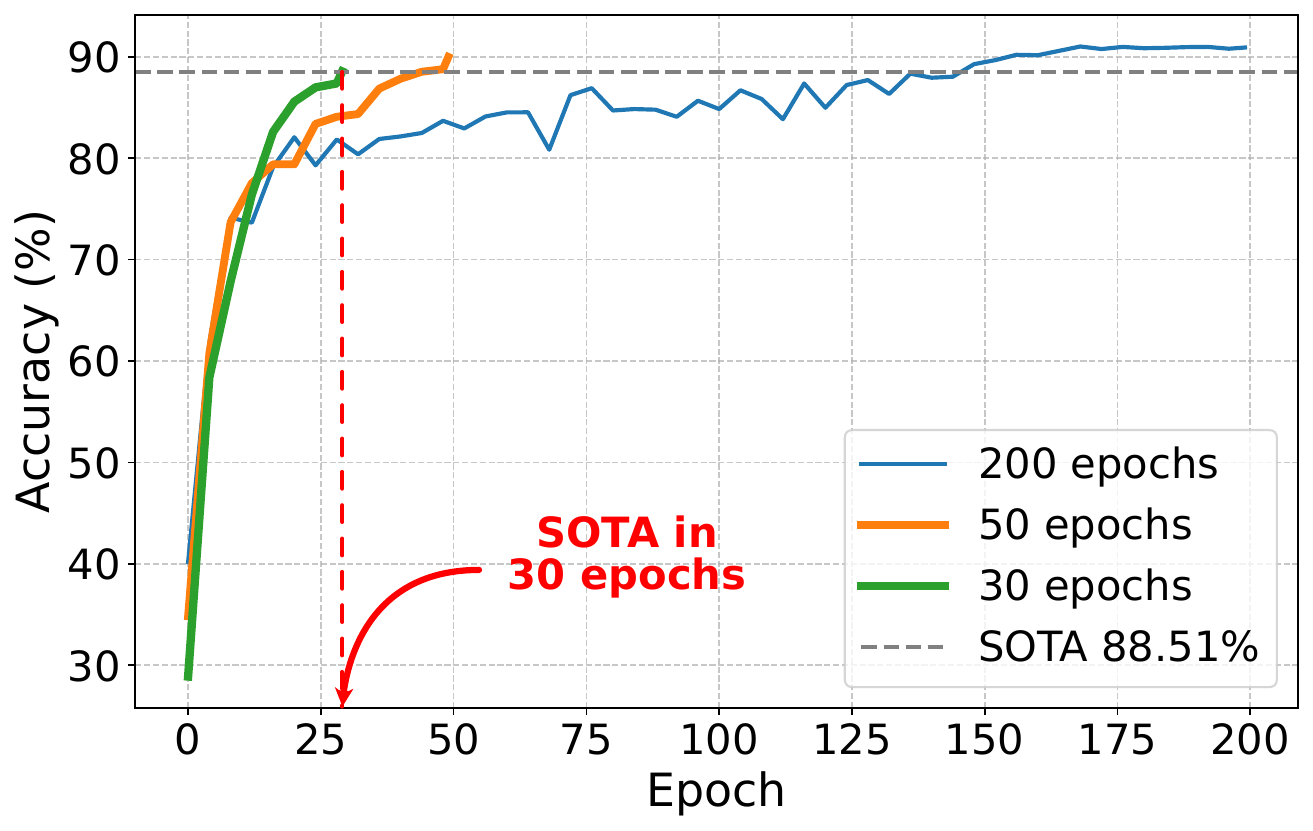}
        \caption{3 Bit (8 colors).}
    \end{subfigure}
    \caption{Illustration of the test accuracy of our method using ResNet-18 on CIFAR-10.}
    \label{fig:train_accelerate}
\end{figure}
\vspace{-1.5em}

\subsection{Additional Analysis}
\textbf{Training Efficiency.} Figure \ref{fig:train_accelerate} illustrates a key limitation of traditional dataset pruning. Although pruning reduces the dataset from $N$ to $\alpha N$ samples ($\alpha < 1$), the total number of parameter updates remains nearly unchanged to ensure convergence. With batch size $b$, the original dataset requires $N/b$ updates per epoch, while the pruned dataset requires $N/(\alpha b)$ to match optimization steps. This compensatory mechanism cancels potential computational gains, so reduced dataset size does not improve training efficiency. Hence, our comparison at the same compression ratio in Table \ref{tab:pruning_methods_combined} is fair.
\vspace{-1.0em} 
\begin{table}[H]
    \centering
    \scriptsize

    \begin{adjustbox}{max width=\linewidth}
    \begin{tabular}{cc}
    
        % ---------- Left Table ----------
        \begin{minipage}[t]{0.46\textwidth}
            \centering
            \caption{Combining color quantization and dataset pruning for extreme compression ratio. The pruning rate is 90\%, and color quantization is $q$-bit.}
            \setlength{\tabcolsep}{0.15em}
            \resizebox{1.0\linewidth}{!}{
            \begin{tabular}{l|cccc|cccc}
                \toprule
                & \multicolumn{4}{c|}{CIFAR-10} & \multicolumn{4}{c}{CIFAR-100} \\
                Color Bits & $q=5$ & $q=4$ & $q=3$ & $q=2$ & $q=5$ & $q=4$ & $q=3$ & $q=2$ \\
                Total Ratio & 98.0\% & 98.4\% & 98.7\% & 99.2\% & 98.0\% & 98.4\% & 98.7\% & 99.2\% \\
                \midrule
                Random & 55.13 & 50.48 & 46.19 & 31.44 & 27.03 & 23.34 & 19.18 & 15.47 \\
                EL2N & 17.11 & 16.36 & 14.69 & 13.11 & 7.01 & 6.94 & 5.38 & 4.33 \\
                CCS$_\mathrm{AUM}$ & 64.20 & 59.69 & 49.35 & 46.25 & 23.14 & 22.58 & 21.53 & 20.11 \\
                TDDS & 63.28 & 55.18 & 46.92 & 43.05 & 22.66 & 21.52 & 20.62 & 18.01 \\
                \textbf{Ours + CCS$_\mathrm{AUM}$} & \textbf{79.92} & \textbf{79.53} & \textbf{76.01} & \textbf{70.73} & \textbf{46.26} & \textbf{40.22} & \textbf{39.49} & \textbf{30.94} \\
                \bottomrule
            \end{tabular}
            \label{tab:pruning_methods_cifar_combined}
            }
        \end{minipage}
 
        &

        % ---------- Right Table ----------
        \begin{minipage}[t]{0.45\textwidth}
            \centering
            \caption{Comparison of dataset distillation algorithms and our algorithm on different datasets using ResNet-18. Entries with \gray{OOM} denotes Out of Memory.}
            \setlength{\tabcolsep}{0.1em}
            \resizebox{0.91\linewidth}{!}{
            \begin{tabular}{l|cc|cc|cc}
                \toprule
                & \multicolumn{2}{c|}{CIFAR-10} & \multicolumn{2}{c|}{Tiny-ImageNet} & \multicolumn{2}{c}{ImageNet-1K} \\
                & IPC=50 & IPC=10 & IPC=100 & IPC=50 & IPC=100 & IPC=50 \\
                \midrule
                DM & 63.0 & 53.9 & \gray{OOM} & \gray{OOM} & \gray{OOM} & \gray{OOM} \\
                MTT & 71.6 & 65.3 & 28.0 & \gray{OOM} & \gray{OOM} & \gray{OOM} \\
                SRe2L & 47.5 & 27.2 & 49.7 & 41.1 & 52.8 & 46.8 \\
                G-VBSM & 59.2 & 53.5 & 51.0 & 47.6 & 55.7 & 51.8 \\
                \textbf{Ours} & \textbf{70.9} & \textbf{67.4} & \textbf{57.6} & \textbf{50.5} & \textbf{59.6} & \textbf{54.4} \\
                \bottomrule
            \end{tabular}}
            \label{tab:Dataset_Distillation}
        \end{minipage}
    \end{tabular}
    \end{adjustbox}

    \label{tab:quant_distill_side_by_side}
\end{table} 
\vspace{-1.2em} 
\textbf{Network Generalization.}
Table \ref{tab:pruning_methods_cifar10_combined} shows that our algorithm outperforms others on larger networks (ResNet-34, ResNet-50), demonstrating strong transferability across architectures. Results for additional networks are provided in Appendix~\ref{appendix:Network}.

\textbf{Combine with Dataset Pruning and Comparison with Dataset Distillation.}
Since our method is orthogonal to dataset pruning, it can be combined with pruning to achieve higher compression. Table~\ref{tab:pruning_methods_cifar_combined} reports results of integrating CCS-based coreset selection (10\% retention; ~\cite{zheng2022coverage}) with $q$-bit color quantization, reaching extreme ratios up to 99.2\% (70.73\% accuracy on CIFAR-10), which highlight the effectiveness of hybrid strategies for extreme data compression.
For  dataset distillation, we use G-VBSM~\citep{shao2024generalized}, SRe2L~\citep{shao2024generalized}, MTT~\citep{cazenavette2022dataset}, and DM~\citep{cazenavette2022dataset} as baselines. To match the compression rates in~\cite{shao2024generalized}, we apply CCS$_\mathrm{AUM}$ with a pruning ratio \(r\%\) and quantize the coreset to \(q\) bits. For CIFAR-10 (IPC = 10, 99.8\% compression; IPC = 50, 99\%), we set \((r\%, q)\) to (95\%, 1) and (75\%, 1). For Tiny-ImageNet (IPC = 50, 90\%; IPC = 100, 80\%), we set \((r\%, q)\) to (0\%, 5) and (0\%, 2). For ImageNet-1K (IPC = 50, 96.1\%; IPC = 100, 92.1\%), we set \((r\%, q)\) to (50\%, 2) and (60\%, 5). This maintains consistency with the original paper. Table~\ref{tab:Dataset_Distillation} shows our algorithm outperforms others.

\textbf{Additional Analysis and Visualization.} 
Additional analysis is provided in Appendix~\ref{appendix:additional-analysis}, and visualizations of all datasets are provided in the Appendix~\ref{appendix:visualization}.

\section{Conclusion}
We present a novel dataset condensation approach that leverages color quantization to reduce storage, improving data efficiency and enabling new forms of compact data representation. While effective, our work has several limitations. Future research could investigate adaptive, per-image quantization strategies to balance compression and accuracy more flexibly. In addition, developing neural architectures specifically optimized for color-quantized data, rather than full-color inputs, may further enhance performance on compressed datasets. These directions highlight the potential of integrating quantization with dataset-level learning to advance efficient and robust deep learning.

\section*{Acknowledgement}
This research is supported by A*STAR Career Development Fund (CDF) under Grant C243512011, the National Research Foundation, Singapore under its National Large Language Models Funding Initiative (AISG Award No: AISG-NMLP-2024-003). Any opinions, findings and conclusions or recommendations expressed in this material are those of the author(s) and do not reflect the views of National Research Foundation, Singapore.

\bibliography{iclr2026_conference}
\bibliographystyle{iclr2026_conference}

\appendix

\section{Experiments Setup}
\subsection{Datasets}
\label{appendix:dataset}

\textbf{CIFAR-10/CIFAR-100} are datasets that consist of 10 and 100 classes, respectively, with image resolutions of $32 \times 32$ pixels.
We use a ResNet-18 architecture~\citep{he2016deep} to train the models for 40,000 iterations with a batch size of 256, equivalent to approximately 200 epochs. Optimization is performed using SGD with a momentum of 0.9, a weight decay of 0.0002, and an initial learning rate of 0.1. A cosine annealing scheduler~\citep{loshchilov2016sgdr} is applied with a minimum learning rate of 0.0001. Data augmentation includes a 4-pixel padding crop and random horizontal flips. The experimental setup follows that of~\citep{zheng2022coverage}.

\textbf{Tiny-ImageNet-200} is a subset of the ImageNet-1K dataset~\citep{le2015tiny}, containing 200 classes. Each class consists of 500 medium-resolution images, and the spatial size of the images is $64 \times 64$ pixels.
A ResNet-18 architecture~\citep{he2016deep} is used, and models are trained for 60 epochs with a batch size of 128. Optimization is performed using SGD with a momentum of 0.9, a weight decay of 0.0002, and an initial learning rate of 0.1. A cosine annealing scheduler~\citep{loshchilov2016sgdr} is applied with a minimum learning rate of 0.0001. Data augmentation includes an 8-pixel padding crop and random horizontal flips. The experimental setup follows that of~\citep{hou2020learning}.

\textbf{ImageNet-1K} is a dataset containing 1,000 classes and 1,281,167 images in total. The images are resized to a resolution of $224 \times 224$ pixels for training.
ResNet-34~\citep{he2016deep} is adopted as the network architecture. Models are trained for 300,000 iterations with a batch size of 256, corresponding to approximately 60 epochs. Optimization uses SGD with a momentum of 0.9, a weight decay of 0.0001, and an initial learning rate of 0.1. A cosine annealing learning rate scheduler is employed. The experimental setup follows that of~\citep{zheng2022coverage}.
\subsection{Baselines}
\subsubsection{Dataset Pruning}
\textbf{Random} randomly selects partial data from the full dataset to form a coreset.

\textbf{Entropy}~\citep{paul2021deep} quantifies sample uncertainty. Samples with higher entropy are considered to have a greater influence on model optimization. The entropy is computed using predicted probabilities at the conclusion of training.

\textbf{Forgetting}~\citep{toneva2018empirical} measures the frequency of forgetting events during training. Samples that are consistently remembered can be removed with minimal impact on performance.

\textbf{CCS}~\citep{zheng2022coverage} applies stratified sampling variations based on importance scores to enhance coreset coverage. This method can also integrate other criteria to further improve results.

\textbf{AUM}~\citep{pleiss2020identifying} selects samples with the highest area under the margin. The margin is defined as the probability gap between the target class and the next largest class across all training epochs. A larger AUM suggests that a sample is of higher importance.

\textbf{EL2N}~\citep{paul2021deep} selects samples based on larger gradient magnitudes, which can be effectively approximated by error vector scores. For practical evaluation, only the average of the first 10 epochs' error vector scores is considered.

\textbf{TDDS}~\citep{zhang2024spanning} first captures each sample's contribution throughout training to integrate detailed dynamics. It then analyzes the variability of these contributions to identify well-generalized samples.

\subsubsection{Color Quantization}

\textbf{ColorCNN}~\citep{zhang2024spanning} preserves critical image structures and produces a quantized version with similar class activation maps to the original image. During training, non-differentiable parts are approximated, and regularization is used to maintain similarity and prevent premature convergence. The ColorCNN is trained end-to-end with classification loss, incorporating color jittering to improve robustness.

\textbf{ColorCNN+}~\citep{hou2024scalable} is an extended model that supports multiple color space sizes by reducing feature dimensions from a predefined number $D$ to the desired color space size $C$ using channel-wise average pooling. Unlike the original ColorCNN, which adapts the network structure based on the color space size.

\textbf{CQFormer}~\citep{su2023name} is a model designed for color quantization using a dual-branch architecture. The annotation branch processes feature activation using a UNeXt network, while the palette branch generates color palettes through a Palette Attention Module (PAM) with reference queries. The model operates across training and testing stages, incorporating intra-cluster color similarity regularization to enhance color classification and detection capabilities by learning a compact color representation.

\textbf{MedianCut}~\citep{heckbert1982color} is a technique used to reduce the number of colors in an image while preserving its overall visual appearance. The algorithm works by recursively dividing the color space into smaller regions (bins) based on the distribution of pixel colors.

\begin{table}[ht]
    \centering
    \small
    \setlength{\tabcolsep}{4pt}
    \renewcommand{\arraystretch}{1.4}   
    \resizebox{0.9\textwidth}{!}{
    \begin{tabular}{l|cccccc|cccccc|cccccc}
        \toprule
        & \multicolumn{6}{c|}{CIFAR-10} & \multicolumn{6}{c|}{CIFAR-100} & \multicolumn{6}{c}{Tiny-ImageNet} \\
        \cmidrule{2-19}
        & \multicolumn{6}{c|}{Color Bits} & \multicolumn{6}{c|}{Color Bits} & \multicolumn{6}{c}{Color Bits} \\
        & 1 & 2 & 3 & 4 & 5 & 6 & 1 & 2 & 3 & 4 & 5 & 6 & 1 & 2 & 3 & 4 & 5 & 6 \\
        \midrule
        OCTree & 57.91 & 68.95 & 75.12 & 77.24 & 80.29 & 84.99 & 27.93 & 30.15 & 39.02 & 43.34 & 47.39 & 52.89 & 24.63 & 26.89 & 34.78 & 40.12 & 44.16 & 49.45 \\
        & {\footnotesize ±0.52} & {\footnotesize ±0.63} & {\footnotesize ±0.71} & {\footnotesize ±0.68} & {\footnotesize ±0.75} & {\footnotesize ±0.25} & {\footnotesize ±0.42} & {\footnotesize ±0.45} & {\footnotesize ±0.52} & {\footnotesize ±0.56} & {\footnotesize ±0.61} & {\footnotesize ±0.65} & {\footnotesize ±0.38} & {\footnotesize ±0.41} & {\footnotesize ±0.48} & {\footnotesize ±0.53} & {\footnotesize ±0.57} & {\footnotesize ±0.62} \\
        CQFormer & 49.52 & 52.53 & 62.91 & 71.34 & 75.49 & 79.89 & 19.92 & 20.01 & 29.95 & 32.81 & 39.62 & 41.89 & 19.91 & 20.01 & 29.95 & 32.81 & 39.65 & 41.89 \\
        & {\footnotesize ±0.48} & {\footnotesize ±0.51} & {\footnotesize ±0.58} & {\footnotesize ±0.65} & {\footnotesize ±0.69} & {\footnotesize ±0.73} & {\footnotesize ±0.35} & {\footnotesize ±0.36} & {\footnotesize ±0.45} & {\footnotesize ±0.47} & {\footnotesize ±0.52} & {\footnotesize ±0.54} & {\footnotesize ±0.34} & {\footnotesize ±0.35} & {\footnotesize ±0.44} & {\footnotesize ±0.46} & {\footnotesize ±0.51} & {\footnotesize ±0.53} \\
        ColorCNN+ & 48.77 & 49.66 & 59.92 & 68.34 & 72.59 & 82.89 & 18.53 & 19.52 & 27.92 & 29.34 & 36.49 & 40.99 & 15.21 & 16.38 & 24.53 & 26.12 & 33.25 & 37.55 \\
        & {\footnotesize ±0.47} & {\footnotesize ±0.48} & {\footnotesize ±0.56} & {\footnotesize ±0.63} & {\footnotesize ±0.67} & {\footnotesize ±0.75} & {\footnotesize ±0.33} & {\footnotesize ±0.34} & {\footnotesize ±0.42} & {\footnotesize ±0.44} & {\footnotesize ±0.49} & {\footnotesize ±0.52} & {\footnotesize ±0.31} & {\footnotesize ±0.32} & {\footnotesize ±0.38} & {\footnotesize ±0.41} & {\footnotesize ±0.46} & {\footnotesize ±0.49} \\
        ColorCNN & 44.12 & 59.48 & 70.23 & 80.34 & 84.49 & 86.89 & 16.54 & 22.81 & 34.93 & 38.64 & 40.19 & 45.39 & 13.28 & 19.42 & 31.63 & 35.21 & 36.89 & 42.11 \\
        & {\footnotesize ±0.45} & {\footnotesize ±0.55} & {\footnotesize ±0.64} & {\footnotesize ±0.72} & {\footnotesize ±0.77} & {\footnotesize ±0.79} & {\footnotesize ±0.32} & {\footnotesize ±0.37} & {\footnotesize ±0.48} & {\footnotesize ±0.51} & {\footnotesize ±0.53} & {\footnotesize ±0.57} & {\footnotesize ±0.29} & {\footnotesize ±0.34} & {\footnotesize ±0.45} & {\footnotesize ±0.48} & {\footnotesize ±0.49} & {\footnotesize ±0.54} \\
        MedianCut & 70.91 & 78.15 & 81.02 & 84.34 & 86.39 & 88.89 & 29.96 & 35.15 & 42.02 & 49.34 & 51.39 & 55.89 & 26.58 & 31.87 & 38.74 & 46.12 & 48.15 & 52.56 \\
        & {\footnotesize ±0.65} & {\footnotesize ±0.71} & {\footnotesize ±0.74} & {\footnotesize ±0.76} & {\footnotesize ±0.78} & {\footnotesize ±0.81} & {\footnotesize ±0.44} & {\footnotesize ±0.48} & {\footnotesize ±0.54} & {\footnotesize ±0.61} & {\footnotesize ±0.63} & {\footnotesize ±0.67} & {\footnotesize ±0.41} & {\footnotesize ±0.46} & {\footnotesize ±0.51} & {\footnotesize ±0.58} & {\footnotesize ±0.59} & {\footnotesize ±0.64} \\
        \textbf{Ours} & \textbf{79.94} & \textbf{89.15} & \textbf{91.02} & \textbf{93.34} & \textbf{94.39} & \textbf{94.89} & \textbf{38.44} & \textbf{57.69} & \textbf{65.02} & \textbf{68.09} & \textbf{70.31} & \textbf{71.55} & \textbf{36.44} & \textbf{50.51} & \textbf{52.12} & \textbf{55.09} & \textbf{57.61} & \textbf{60.02} \\
        & {\footnotesize ±0.72} & {\footnotesize ±0.33} & {\footnotesize ±0.84} & {\footnotesize ±0.86} & {\footnotesize ±0.87} & {\footnotesize ±0.88} & {\footnotesize ±0.51} & {\footnotesize ±0.67} & {\footnotesize ±0.73} & {\footnotesize ±0.76} & {\footnotesize ±0.78} & {\footnotesize ±0.79} & {\footnotesize ±0.49} & {\footnotesize ±0.62} & {\footnotesize ±0.64} & {\footnotesize ±0.66} & {\footnotesize ±0.68} & {\footnotesize ±0.71} \\
        \bottomrule
    \end{tabular}}
    \caption{Performance comparison for different methods across varying color bits on CIFAR-10, CIFAR-100, and Tiny-ImageNet datasets.
    Mean and standard deviation are computed from 5 independent runs.}
     
    \label{tab:comparison_all}
\end{table}
\textbf{OCTree}~\citep{gervautz1988simple} builds an octree data structure where each node represents a color or a group of similar colors. Starting with all colors of an image in the octree, the process iteratively merges nodes (colors) that are closest together until a target number of colors is reached.
\label{appendix:baseline}

% Figure \ref{fig:quantization-comparison} and Figure \ref{fig:Tiny-ImageNet} present qualitative results comparing different color quantization ratios on the CIFAR-10/100 train-set and the Tiny-ImageNet dataset.
All images are randomly sampled from the train-set to ensure representativeness. The visualization demonstrates how our method progressively reduces the color palette while maintaining the visual fidelity of the original images. As the quantization ratio increases (i.e., fewer color bits), we observe that our approach effectively preserves key structural details and salient features, even under aggressive color reduction scenarios. 

\section{Additional Information}
\label{appendix:primary-cq}
\subsection{Primary Table for Color Quantization}
Table \ref{tab:comparison_all} shows the results of color quantization on different datasets. We ran each experiment five times to calculate the mean and standard deviation. The experimental results demonstrate that our color quantization algorithm outperforms other color quantization algorithms in improving the performance of the train-set after color quantization.

\section{Additional Experiments}

\subsection{Ablation of Sobel Operator}
Table \ref{tab:sobel} demonstrates that incorporating edge distribution minimization consistently improves the color quantization performance across all bit depths (1-5) on CIFAR-10. Notably, we observe performance gains ranging from 2.26\% to 3.10\%, with the most significant improvement occurring at 1-bit quantization (76.8\% → 79.90\%). These results validate that preserving edge information during quantization leads to better overall performance.
\label{appendix:sobel}

\begin{table}[H]
    \centering
    \footnotesize
    \caption{Effect of using minimizing the edge distribution differences on CIFAR-10.}
    \begin{tabular}{cccccc}
        \toprule
        \textbf{Color bit} & \textbf{1} & \textbf{2} & \textbf{3} & \textbf{4} & \textbf{5}\\
        \midrule
        No using & 76.80 & 87.05 & 89.22 & 91.33 & 92.13\\
        \textbf{using} & \textbf{79.90} & \textbf{89.15} & \textbf{91.02} & \textbf{93.34} & \textbf{94.39}\\
        \bottomrule
    \end{tabular}
    \label{tab:sobel}
\end{table}

\subsection{Generalization to Different Networks}
Table \ref{tab:pruning_methods_cifar10_sh} presents a comparative analysis of our proposed algorithm against existing dataset pruning methods across Shufflenet~\citep{zhang2018shufflenet} and MobileNet-v2~\citep{sandler2018mobilenetv2} architectures. The empirical results demonstrate that our approach consistently outperforms baseline methods across different compression ratios, achieving superior accuracy on both network architectures.

\label{appendix:Network}

\begin{table}[H]
    \centering
    \footnotesize
    \setlength{\tabcolsep}{0.8em}
    \caption{Comparison of dataset pruning algorithms and our algorithm on CIFAR-10 with ShufffleNet and MobileNet-v2.}
    \begin{tabular}{l|ccccc|ccccc}
        \toprule
        & \multicolumn{5}{c|}{ShufffleNet, CIFAR-10} & \multicolumn{5}{c}{MobileNet-v2, CIFAR-10} \\
        & 80\% & 83\% & 87.5\% & 92\% & 96\% & 80\% & 83\% & 87.5\% & 92\% & 96\% \\
        \midrule
        Random & 83.05 & 80.29 & 72.18 & 70.34 & 60.18 & 81.25 & 78.09 & 70.38 & 68.04 & 57.88 \\
        EL2N & 69.20 & 65.32 & 23.68 & 22.13 & 20.19 & 67.20 & 63.02 & 21.88 & 20.13 & 18.19 \\
        AUM & 54.64 & 45.49 & 27.78 & 20.62 & 17.35 & 55.84 & 46.59 & 27.78 & 23.30 & 19.05 \\
        CCS$_\mathrm{AUM}$ & 82.14 & 80.32 & 77.51 & 65.94 & 63.45 & 80.64 & 78.02 & 75.01 & 63.44 & 61.15 \\
        TDDS & 83.08 & 81.65 & 80.42 & 66.55 & 62.98 & 81.28 & 79.15 & 77.92 & 65.05 & 59.48 \\
        \textbf{Ours} & \textbf{89.9} & \textbf{87.98} & \textbf{86.31} & \textbf{83.67} & \textbf{74.41} & \textbf{91.19} & \textbf{88.84} & \textbf{89.25} & \textbf{85.17} & \textbf{76.26} \\
        \bottomrule
    \end{tabular}
    \label{tab:pruning_methods_cifar10_sh}
\end{table}

Table \ref{tab:pruning_methods_cifar100_Transformer} shows a comparative analysis of our proposed algorithm against existing dataset pruning methods across Swin Transformer~\citep{liu2021swin} and ViT-Small~\citep{lee2021vision}.

\begin{table}[H]
    \centering
    \footnotesize
    \setlength{\tabcolsep}{0.8em}
    \caption{Comparison of dataset pruning algorithms and our algorithm on Transformer-Based Model. For CIFAR-10, we tested our algorithms on Swin Transformer; For CIFAR-100, we tested our algorithms on ViT-Small.}
    \begin{tabular}{l|ccccc|ccccc}
        \toprule
        & \multicolumn{5}{c|}{CIFAR-10} & \multicolumn{5}{c}{CIFAR-100} \\
        & 80\% & 83\% & 87.5\% & 92\% & 96\% & 80\% & 83\% & 87.5\% & 92\% & 96\% \\
        \midrule
        Random & 82.26 & 79.18 & 77.09 & 69.27 & 64.18   & 33.83 & 30.96 & 26.19 & 25.61 & 22.68 \\
        EL2N & 59.13 & 55.15 & 27.77 & 24.97 & 21.64   & 22.49 & 14.55 & 11.67 & 9.18 & 8.76 \\
        AUM & 65.61 & 56.51 & 28.65 & 23.22 & 20.17       & 22.58 & 15.32 & 12.88 & 10.15 & 6.08 \\
        CCS$_\mathrm{AUM}$ & 84.38 & 80.82 & 75.18 & 71.22 & 67.01    & 59.18 & 55.35 & 51.16 & 27.13 & 22.84 \\
        \textbf{Ours} & \textbf{87.66} & \textbf{84.91} & \textbf{80.82} & \textbf{76.83} & \textbf{70.54}     & \textbf{68.38} & \textbf{65.19} & \textbf{63.72} & \textbf{47.19} & \textbf{35.42} \\
        \bottomrule
    \end{tabular}
    \label{tab:pruning_methods_cifar100_Transformer}
\end{table} 

\subsection{Ablation of Feature Extraction Algorithms}

During the  Cluster for Color Perceptual Consistency step, we used a pre-trained model to extract features. Table \ref{tab:cluster_features_Stra} compares the clustering performance of different features, including PCA, Gray Level Co-occurrence Matrix (GLCM), Color Coherence Vector (CCV), RGB Color Histograms, and Shallow-layer Feature Maps. The results show that clustering with Shallow-layer Feature Maps achieves the best performance.

\begin{table}[H]
    \centering
    \footnotesize
    \caption{Different feature extraction strategies on CIFAR-10 trained on ResNet-18.}
    \begin{tabular}{l|ccccc}
        \toprule
        \textbf{Color Bits} & \textbf{1} & \textbf{2} & \textbf{3} & \textbf{4} & \textbf{5} \\
        \midrule
        PCA & 50.11 & 61.78 & 70.23 & 73.45 & 78.47 \\
        Gray Level Co-occurrence Matrix & 70.15 & 83.17 & 85.49 & 86.17 & 90.75 \\
        Color Coherence Vector & 74.15 & 86.97 & 87.19 & 89.77 & 92.35 \\
        RGB Color Histograms & 75.95 & 86.18 & 88.49 & 90.17 & 91.45 \\
        \textbf{Shallow-layer Feature Map} & \textbf{79.90} & \textbf{89.15} & \textbf{91.02} & \textbf{93.34} & \textbf{94.39} \\
        \bottomrule
    \end{tabular}
    \label{tab:cluster_features_Stra}
\end{table}

Table \ref{tab:layer} presents the results of using different layers for feature extraction. Specifically, we evaluate  Shallow-layer feature maps from the first residual block,  Mid-layer1 feature maps from the second residual block, Mid-layer2 feature maps from the third residual block, and Final-layer feature maps from the final residual block. The results indicate that Shallow-layer feature maps achieve the best performance.
\begin{table}[h]
    \centering
    \footnotesize
    \caption{Comparison of Different layers on CIFAR-10 trained on ResNet-18.}
    \begin{tabular}{l|ccccc}
        \toprule
        \textbf{Color Bits} & \textbf{1} & \textbf{2} & \textbf{3} & \textbf{4} & \textbf{5} \\
        \midrule
        Mid-layer2 feature maps & 69.15 & 79.81 & 81.09 & 83.76 & 87.75 \\
         Mid-layer1 feature maps & 74.15 & 86.97 & 87.19 & 89.77 & 92.35 \\
         Final-layer Feature Map &  42.10  &  53.78  &  66.39  &  75.15  &  80.44  \\
        \textbf{Shallow-layer Feature Map} & \textbf{79.90} & \textbf{89.15} & \textbf{91.02} & \textbf{93.34} & \textbf{94.39} \\
        \bottomrule
    \end{tabular}
    \label{tab:layer}
\end{table} 
\label{appendix:Feature_Extraction_Algorithms}

\subsection{Ablation of Grad-CAM++ Percentage}
Table \ref{tab:Grad_CAM_CIFAR10} presents the performance under different Grad-CAM++ retention percentages. The results indicate that preserving 50\% of the pixels achieves the best performance.
\begin{table}[H]
    \centering
    \footnotesize
    \caption{Utilizing Grad-CAM++ attention maps to maintain the pixel distribution ratio $k_{Gra}\%$ on CIFAR-10 and ResNet-18.}
    \begin{tabular}{l|ccccc}
        \toprule
        \textbf{Color Bits} & \textbf{1} & \textbf{2} & \textbf{3} & \textbf{4} & \textbf{5} \\
        \midrule
        80\% & 75.95 & 87.17 & 88.11 & 89.98 & 93.05 \\
        20\% & 77.31 & 87.79 & 88.23 & 90.55 & 91.47 \\
        70\% & 76.15 & 86.97 & 87.41 & 90.99 & 92.55 \\
        60\% & 76.85 & 87.17 & 88.89 & 91.97 & 93.01 \\
        30\% & 77.95 & 88.18 & 89.19 & 92.17 & 93.45 \\
        \textbf{50\%} & \textbf{79.90} & \textbf{89.15} & \textbf{91.02} & \textbf{93.34} & \textbf{94.39} \\
        \bottomrule
    \end{tabular}
    \label{tab:Grad_CAM_CIFAR10}
\end{table}

\label{appendix:Grad_CAM}

\section{Additional Analysis}
\label{appendix:additional-analysis}

\subsection{Experiments in Other Tasks}
Table \ref{tab:image_segmentation} and Table \ref{tab:object_detection} show that our algorithm outperforms other dataset pruning algorithms on both image segmentation and object detection tasks under high compression ratios.

\begin{table}[H]
\centering
\footnotesize
\caption{Comparison of different dataset pruning algorithms on MS COCO.}
\begin{tabular}{cc}
\begin{subtable}[b]{0.48\linewidth}
\centering
\caption{Image Segmentation (MS COCO), Full dataset AP$_{50}$ = 55.20\%.}
\label{tab:image_segmentation}
\begin{tabular}{lcccc}
\toprule
\textbf{Method} & \textbf{80\%} & \textbf{87.5\%} & \textbf{92\%} & \textbf{96\%} \\
\midrule
Random  & 45.60 & 42.07 & 38.15 & 25.99 \\
CCS     & 43.68 & 39.91 & 35.58 & 26.46 \\
EL2N    & 38.98 & 36.12 & 29.91 & 24.26 \\
\textbf{Ours} & \textbf{47.38} & \textbf{44.29} & \textbf{41.68} & \textbf{28.77} \\
\bottomrule
\end{tabular}
\end{subtable}
&
\begin{subtable}[b]{0.48\linewidth}
\centering
\caption{Object Detection (MS COCO), Full dataset AP$_{50}^{\mathrm{bb}}$ = 58.30\%.}
\label{tab:object_detection}
\begin{tabular}{lcccc}
\toprule
\textbf{Method} & \textbf{80\%} & \textbf{87.5\%} & \textbf{92\%} & \textbf{96\%} \\
\midrule
Random  & 48.80 & 44.07 & 40.15 & 31.89 \\
CCS     & 46.71 & 42.89 & 37.08 & 30.15 \\
EL2N    & 40.05 & 39.15 & 31.48 & 27.36 \\
\textbf{Ours} & \textbf{50.78} & \textbf{46.29} & \textbf{42.08} & \textbf{33.37} \\
\bottomrule
\end{tabular}
\end{subtable}
\\
\end{tabular}
\label{tab:coco_pruning}
\end{table}

Table \ref{tab:cifar10_psld_fid} shows that our algorithm outperforms other dataset pruning algorithms on generative tasks.

\begin{table}[H]
\centering
\footnotesize
\caption{Comparison of dataset pruning algorithms on the CIFAR-10 PSLD generative task~\citep{pandey2023complete}. Results are evaluated by FID ($\downarrow$), where lower values indicate better performance.}
\label{tab:cifar10_psld_fid}
\begin{tabular}{lcccc}
\toprule
\textbf{Method} & \textbf{80\%} & \textbf{87.5\%} & \textbf{92\%} & \textbf{96\%} \\
\midrule
Random  & 16.8  & 39.4  & 50.2  & 119.8 \\
CCS     & 15.7  & 37.8  & 52.8  & 121.1 \\
EL2N    & 18.1  & 45.1 & 61.4  & 138.3 \\
\textbf{Ours} & \textbf{11.9} & \textbf{21.3} & \textbf{42.2} & \textbf{113.5} \\
\bottomrule
\end{tabular}
\end{table} 

\subsection{Comparison with Other Tasks}
Table \ref{tab:cifar10_compression_DQ}, Table \ref{tab:imagenet_compression_DQ} and Table \ref{tab:comparison_autopalette} shows comparison with other dataset compression algorithms like DUAL~\citep{cho2025lightweight}, DQ~\citep{zhou2023dataset}, DQAL~\citep{zhao2024dataset}, ADQ~\citep{li2025adaptive} and AutoPalette~\citep{yuan2024color}, our algorithms outperform than other dataset pruning algorithms.

\begin{table}[H]
\centering
\footnotesize
\caption{Comparison of our DCQ with different dataset compression algorithms.}
\begin{tabular}{cc}
\begin{subtable}[b]{0.48\linewidth}
\centering
\caption{CIFAR-10 (ResNet-18)}
\label{tab:cifar10_compression_DQ}
\begin{tabular}{lcccc}
\toprule
\textbf{Method} & \textbf{80\%} & \textbf{87.5\%} & \textbf{92\%} & \textbf{96\%} \\
\midrule
ADQ    & 90.40 & 87.51 & 86.32 & 75.99 \\
DQAL   & 90.20 & 87.61 & 85.30 & 75.80 \\
DQ     & 89.40 & 87.70 & 84.25 & 77.92 \\
DUAL   & 91.42 & 88.99 & 86.15 & 74.83 \\
\textbf{Ours}  & \textbf{94.39} & \textbf{91.02} & \textbf{89.15} & \textbf{79.90} \\
\bottomrule
\end{tabular}
\end{subtable}
&
\begin{subtable}[b]{0.48\linewidth}
\centering
\caption{ImageNet-1k (ResNet-34)}
\label{tab:imagenet_compression_DQ}
\begin{tabular}{lcccc}
\toprule
\textbf{Method} & \textbf{80\%} & \textbf{87.5\%} & \textbf{92\%} & \textbf{96\%} \\
\midrule
ADQ    & 65.31 & 60.91 & 47.02 & 32.19 \\
DQAL   & 62.21 & 60.41 & 46.73 & 31.89 \\
DQ     & 64.30 & 60.45 & 47.17 & 32.27 \\
DUAL   & 66.50 & 61.15 & 46.55 & 33.83 \\
\textbf{Ours} & \textbf{66.99} & \textbf{62.02} & \textbf{49.69} & \textbf{35.95} \\
\bottomrule
\end{tabular}
\end{subtable}
\\
\end{tabular}
\label{tab:dataset_compression_summary}
\end{table}

\begin{table}[H]
\centering
\footnotesize
\caption{Comparison between our algorithm and AutoPalette on CIFAR-10 and CIFAR-100 (ConvNetD3)}
\label{tab:comparison_autopalette}
\begin{tabular}{lcccc}
\toprule
& \multicolumn{2}{c}{\textbf{CIFAR-10}} & \multicolumn{2}{c}{\textbf{CIFAR-100}} \\  
\cmidrule(lr){2-3} \cmidrule(lr){4-5}  
& \textbf{IPC=10} & \textbf{IPC=50} & \textbf{IPC=10} & \textbf{IPC=50} \\
\midrule
AutoPalette & 74.3 & 79.4 & 52.6 & 53.3 \\
\textbf{Ours} & \textbf{77.9} & \textbf{82.3} & \textbf{58.9} & \textbf{61.7} \\
\bottomrule
\end{tabular}
\end{table}

Table \ref{tab:augmentation_combination_cifar10} shows that standard data augmentation further improves performance when training on our color-quantized datasets.

\begin{table}[H]
\centering
\footnotesize
\caption{Combining other dataset augmentation algorithms on CIFAR-10 (ResNet-18) under 1-bit quantization.}
\label{tab:augmentation_combination_cifar10}
\begin{tabular}{l|cccc}
\toprule
\textbf{Method} & \textbf{80\%} & \textbf{87.5\%} & \textbf{92\%} & \textbf{96\%} \\
\midrule
Original setting  & 94.39 & 91.02 & 89.15 & 79.90 \\
Original setting + ColorJitter  & 95.02 & 92.44 & 90.45 & 80.95 \\
Original setting + GaussianBlur & 95.47 & 92.95 & 91.47 & 81.75 \\
\bottomrule
\end{tabular}
\vspace{0.5em}  
\end{table}

Table \ref{tab:stanford_cars_pruning} shows that our algorithm outperforms other algorithms on fine-grained recognition tasks.

\begin{table}[H]
\centering
\footnotesize
\caption{Different dataset pruning algorithms on a fine-grained recognition task (Stanford Cars, API-Net~\citep{zhuang2020learning}). Full dataset performance = 95.3\%.} 
\label{tab:stanford_cars_pruning}
\begin{tabular}{l|cccc}
\toprule
\textbf{Method} & \textbf{80\%} & \textbf{87.5\%} & \textbf{92\%} & \textbf{96\%} \\
\midrule
Random    & 85.95 & 71.28 & 68.19 & 64.55 \\
CCS       & 88.98 & 74.17 & 70.07 & 67.17 \\
EL2N      & 67.32 & 23.98 & 16.31 & 14.85 \\
\textbf{Ours} & \textbf{91.16} & \textbf{76.22} & \textbf{72.34} & \textbf{70.17} \\
\bottomrule
\end{tabular}
\end{table}

Table \ref{tab:cifar10_pruning_noise} shows that our algorithm has high robustness under label noise injection.

\begin{table}[H]
\centering
\footnotesize
\caption{Evaluation of dataset pruning algorithms on CIFAR-10 (ResNet-18) with 30\% symmetric random label noise. Full dataset accuracy = 83.06\%.}
\label{tab:cifar10_pruning_noise}
\begin{tabular}{l|cccc}
\toprule
\textbf{Method} & \textbf{80\%} & \textbf{87.5\%} & \textbf{92\%} & \textbf{96\%} \\
\midrule
Random    & 76.51 & 67.18 & 62.05 & 49.36 \\
CCS       & 78.51 & 69.35 & 64.16 & 51.28 \\
TDDS      & 77.44 & 67.99 & 63.66 & 50.85 \\
\textbf{Ours} & \textbf{82.26} & \textbf{73.44} & \textbf{68.34} & \textbf{59.17} \\
\bottomrule
\end{tabular}
\end{table}

Table \ref{tab:cifar10_epochs_1bit_quant} shows that even when using substantially fewer epochs (e.g., 50 epochs = 9,750 iterations), our method still outperforms SOTA pruning methods trained for ~40,000 iterations, demonstrating its strong performance robustness.

\begin{table}[h]
\centering
\caption{Different number of training epochs on CIFAR-10 (ResNet-18) under 1-bit color quantization}
\label{tab:cifar10_epochs_1bit_quant}
\begin{tabular}{ccc}
\toprule
\textbf{Number of Epochs} & \textbf{Number of Training Iterations} & \textbf{Acc (\%)} \\
\midrule
50  & 9,750   & 75.01 \\
100 & 19,500  & 76.19 \\
125 & 24,375  & 77.69 \\
150 & 29,250  & 78.57 \\
200 & 39,000  & 79.90 \\
\bottomrule
\end{tabular}
\end{table}

\subsection{Computational Complexity}
 Table \ref{tab:dataset_stats} reports the computational complexity of DCQ, which scales linearly with the size of the dataset. Table \ref{tab:time_cost_each} reports the time consumption of each stage in our pipeline. 
 
\begin{table}[H]
\centering
\footnotesize
\caption{Dataset statistics including class count, image distribution, total size, and complexity.}
\label{tab:dataset_stats}
 
\setlength{\tabcolsep}{3pt}
 
\begin{tabular}{lcccc}
\toprule
\textbf{Dataset} & \textbf{\#Classes (C)} & \textbf{Images per Class (IPC)} & \textbf{Total Images} & \textbf{Complexity} \\
\midrule
ImageNet-1K   & 1,000  & $\approx$1200 & $\approx$1.2M  & $O(1{,}000 \times 1200 \times \text{BackboneFLOPs})$ \\
ImageNet-21K  & 21,000 & $\approx$700  & $\approx$14.7M & $O(21{,}000 \times 700 \times \text{BackboneFLOPs})$ \\
\bottomrule
\end{tabular}
 
\setlength{\tabcolsep}{6pt}
\end{table}

\begin{table}[H]
\centering
\footnotesize  
\caption{Time cost of generating different datasets with color quantization (sufficient memory).}
\label{tab:time_cost_each}
 
\setlength{\tabcolsep}{2.5pt}
\begin{tabular}{lcccc}
\toprule
\textbf{Dataset} & \textbf{Palette-Learning} & \textbf{Attention-Computation} & \textbf{Differentiable-Refinement} & \textbf{Total Time} \\
\midrule
ImageNet-1K      & 108 mins                  & 16 mins                          & 30 mins                              & 154 mins             \\
CIFAR-10/100     & 20 seconds                & 2 seconds                        & 30 seconds                           & 52 seconds           \\
\bottomrule
\end{tabular}
\end{table}

\section{Visualization}
\label{appendix:visualization}

\begin{figure}[H]
    \centering
    \begin{minipage}[t]{0.48\textwidth}
        \begin{subfigure}[t]{\textwidth}
            \includegraphics[width=\textwidth]{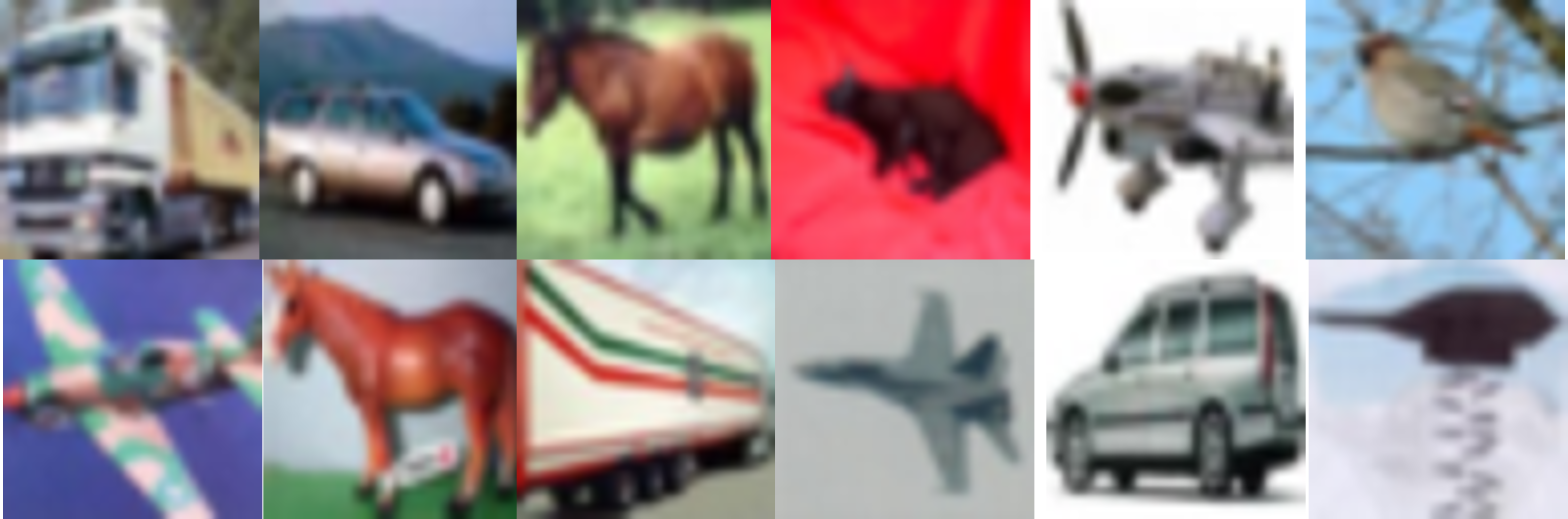}
            \caption{Original (24 Bits, $2^{24}$ colors), CIFAR-10.}
        \end{subfigure}
        
        \vspace{1em}
        
        \begin{subfigure}[t]{\textwidth}
            \includegraphics[width=\textwidth]{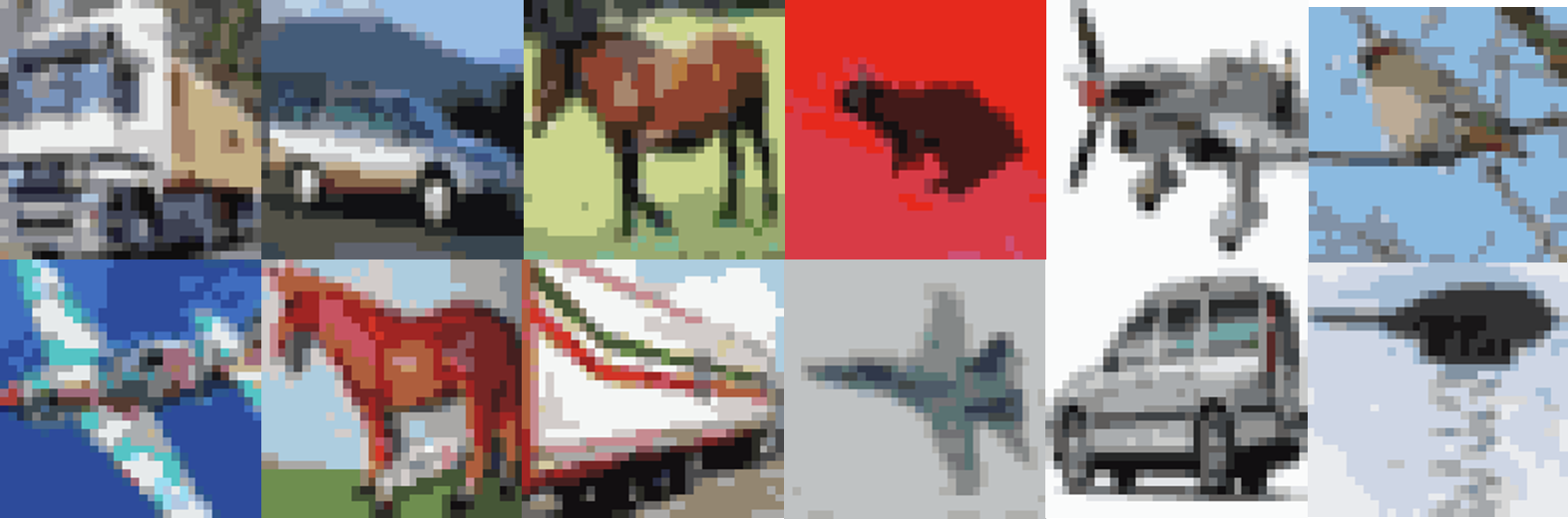}
            \caption{5 Bits (32 colors), CIFAR-10.}
        \end{subfigure}
        
        \vspace{1em}
        
        \begin{subfigure}[t]{\textwidth}
            \includegraphics[width=\textwidth]{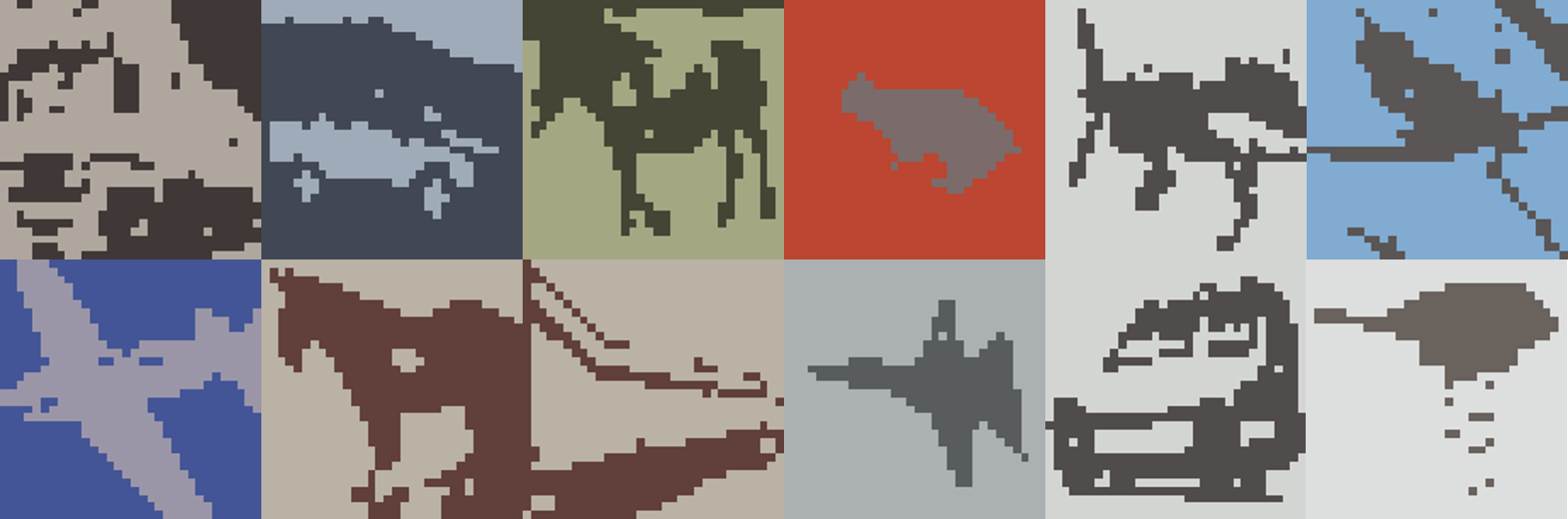}
            \caption{1 Bit (2 colors), CIFAR-10.}
        \end{subfigure}
    \end{minipage}
    \hfill
    \begin{minipage}[t]{0.48\textwidth}
        \begin{subfigure}[t]{\textwidth}
            \includegraphics[width=\textwidth]{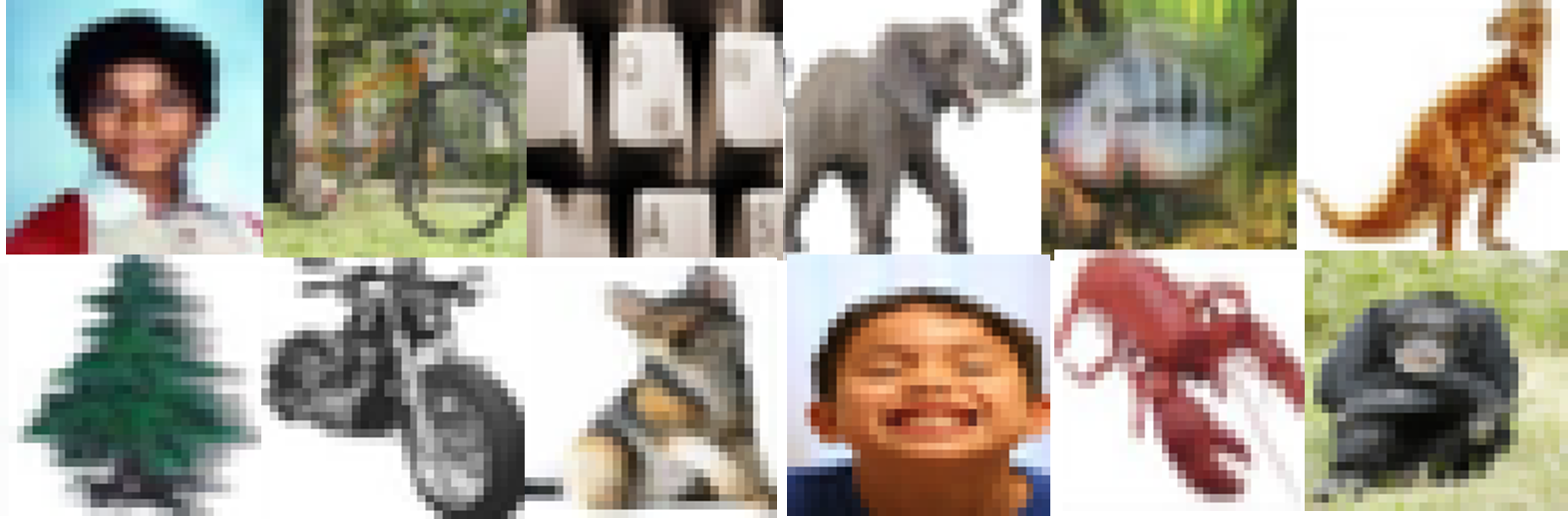}
            \caption{Original (24 Bits, $2^{24}$ colors), CIFAR-100.}
        \end{subfigure}
        
        \vspace{1em}
        
        \begin{subfigure}[t]{\textwidth}
            \includegraphics[width=\textwidth]{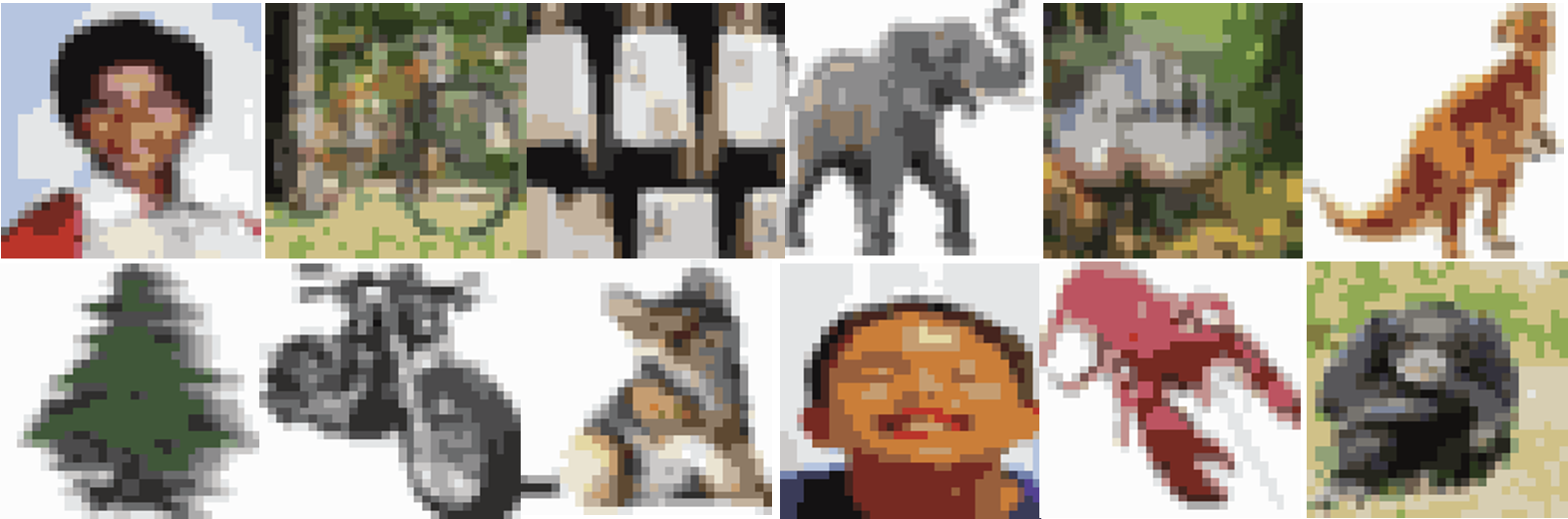}
            \caption{5 Bits (32 colors), CIFAR-100.}
        \end{subfigure}
        
        \vspace{1em}
        
        \begin{subfigure}[t]{\textwidth}
            \includegraphics[width=\textwidth]{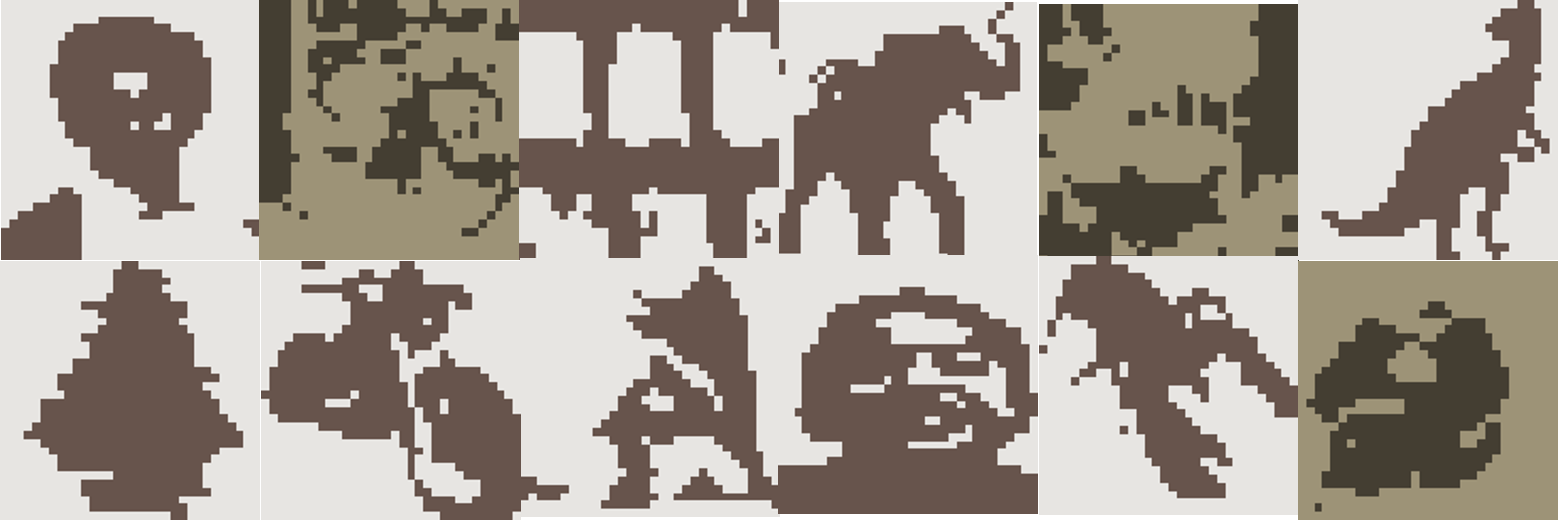}
            \caption{1 Bit (2 colors), CIFAR-100.}
        \end{subfigure}
        
    \end{minipage}

    \caption{Visualization of different datasets under different color quantization ratios. Left column: CIFAR-10 images; Right column: CIFAR-100 images. From top to bottom: original images, 5-bits quantization (32 colors), and 1-bit quantization (2 colors).}
    \label{fig:quantization-comparison}
\end{figure}

\begin{figure}[H]
    \centering
    \begin{subfigure}{\textwidth}
        \centering
        \includegraphics[width=0.9\textwidth]{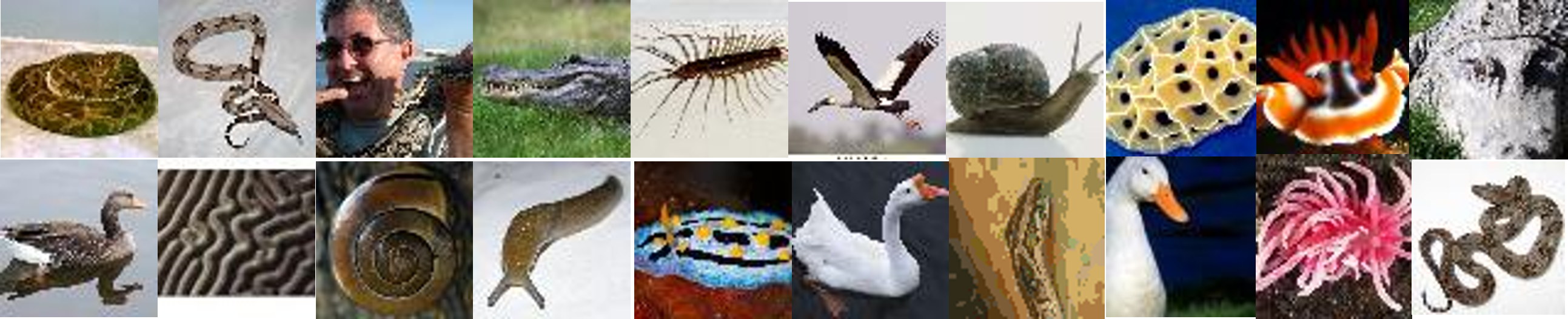}
        \caption{Original (24 Bits, $2^{24}$ colors), Tiny-ImageNet.}
        \label{fig:img1}
    \end{subfigure}
    \vspace{2mm}
    
    \begin{subfigure}{\textwidth}
        \centering
        \includegraphics[width=0.9\textwidth]{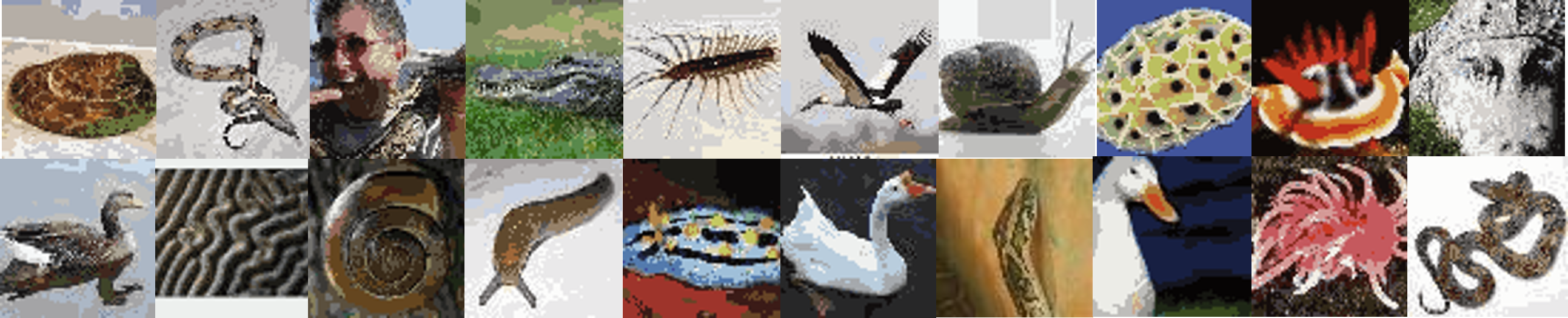}
        \caption{5 Bit (32 colors), Tiny-ImageNet}
        \label{fig:img3}
    \end{subfigure}
    
    \vspace{2mm}
    
    \begin{subfigure}{\textwidth}
        \centering
        \includegraphics[width=0.9\textwidth]{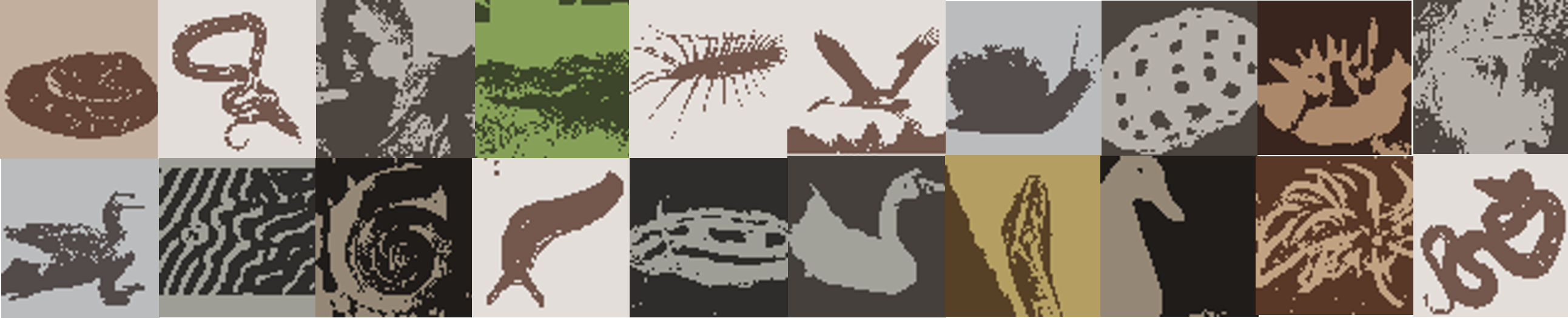}
        \caption{1 Bit (2 colors), Tiny-ImageNet}
        \label{fig:img2}
    \end{subfigure}
    
    \caption{Visualization of Tiny-ImageNet under different color quantization ratios. From top to bottom: original images, 5-bits quantization (32 colors), and 1-bit quantization (2 colors).}
    \label{fig:Tiny-ImageNet}
\end{figure}

Figure~\ref{fig:quantization-comparison} and \ref{fig:Tiny-ImageNet} presents the visualizations of CIFAR-10/100 and Tiny-ImageNet under different color quantization rates, respectively.

\begin{figure}[H]
    \centering
    \begin{minipage}[t]{0.48\textwidth}
        \begin{subfigure}[t]{\textwidth}
            \includegraphics[width=\textwidth]{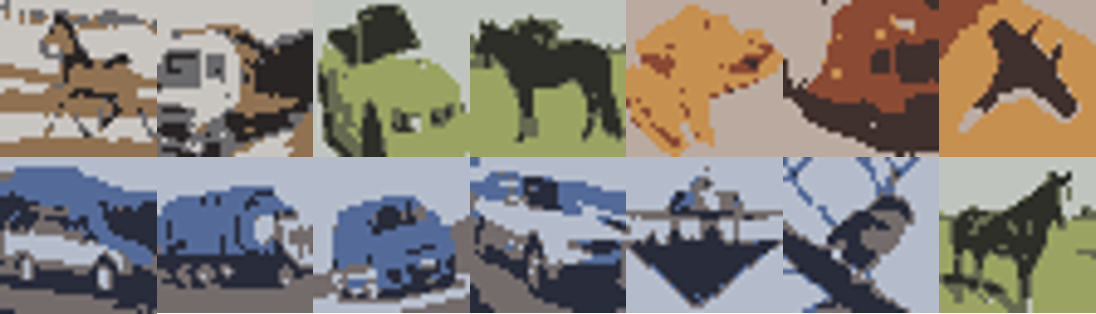}
            \caption{2 Bits (4 colors), CIFAR-10.}
        \end{subfigure}
        
        \vspace{1em}
        
        \begin{subfigure}[t]{\textwidth}
            \includegraphics[width=\textwidth]{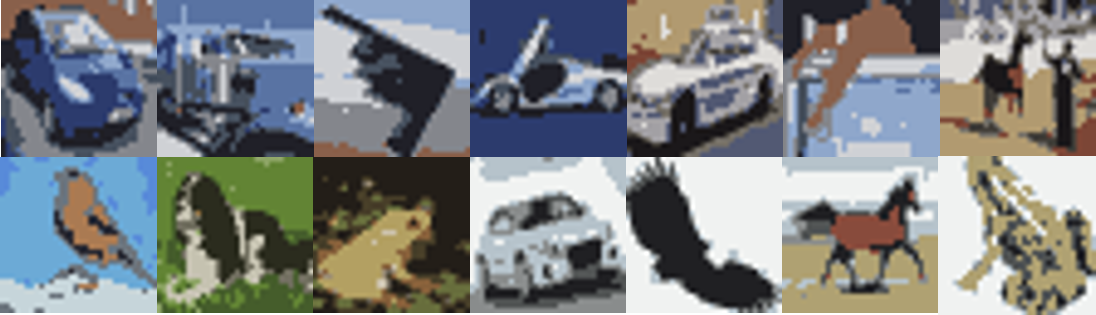}
            \caption{3 Bits (8 colors), CIFAR-10.}
        \end{subfigure}

    \end{minipage}
    \hfill
    \begin{minipage}[t]{0.48\textwidth}
        \begin{subfigure}[t]{\textwidth}
            \includegraphics[width=\textwidth]{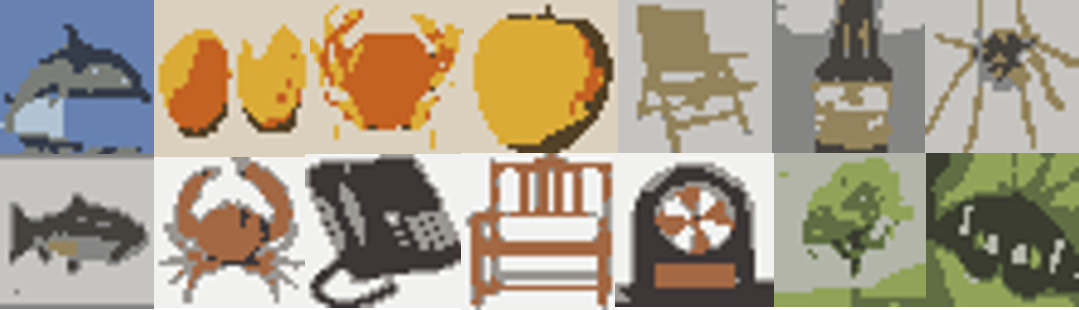}
            \caption{2 Bits (4 colors), CIFAR-100.}
        \end{subfigure}
        
        \vspace{1em}
        
        \begin{subfigure}[t]{\textwidth}
            \includegraphics[width=\textwidth]{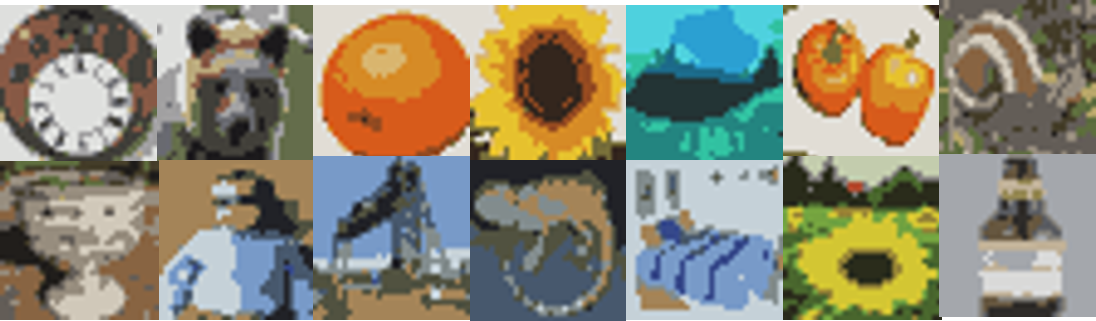}
            \caption{3 Bits (8 colors), CIFAR-100.}
        \end{subfigure}
        
    \end{minipage}

    \caption{Visualization of different datasets under different color quantization ratios. Left column: CIFAR-10 images; Right column: CIFAR-100 images. From top to bottom: 2-bit quantization (4 colors), and 3-bit quantization (8 colors).}
    \label{fig:quantization-comparison-cifar}
\end{figure}

\begin{figure}[H]
    \centering
    \begin{subfigure}{\textwidth}
        \centering
        \includegraphics[width=0.9\textwidth]{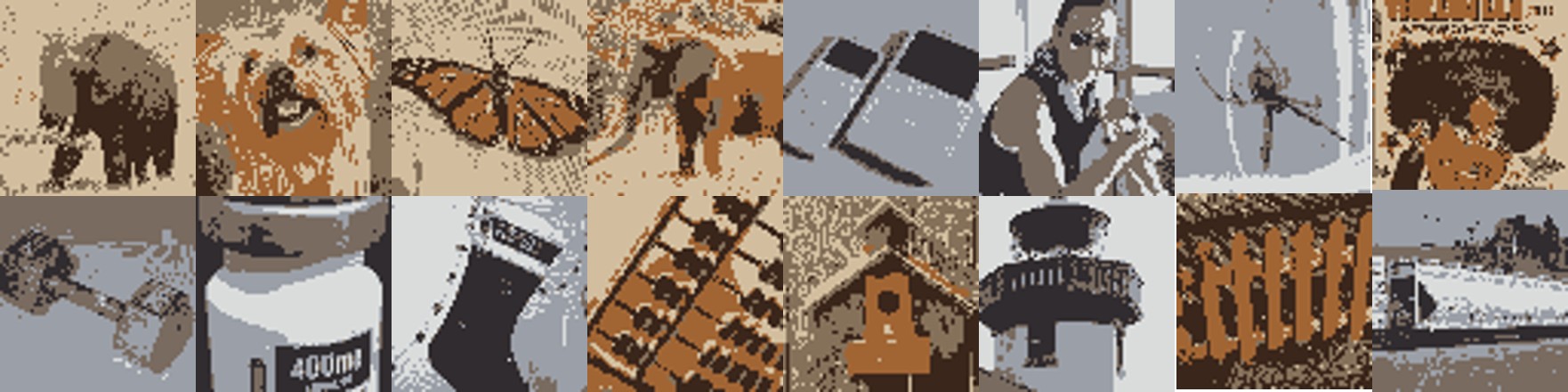}
        \caption{Original (2 Bit (4 colors), Tiny-ImageNet.}
        \label{fig:img1}
    \end{subfigure}
    \vspace{2mm}
    
    \begin{subfigure}{\textwidth}
        \centering
        \includegraphics[width=0.9\textwidth]{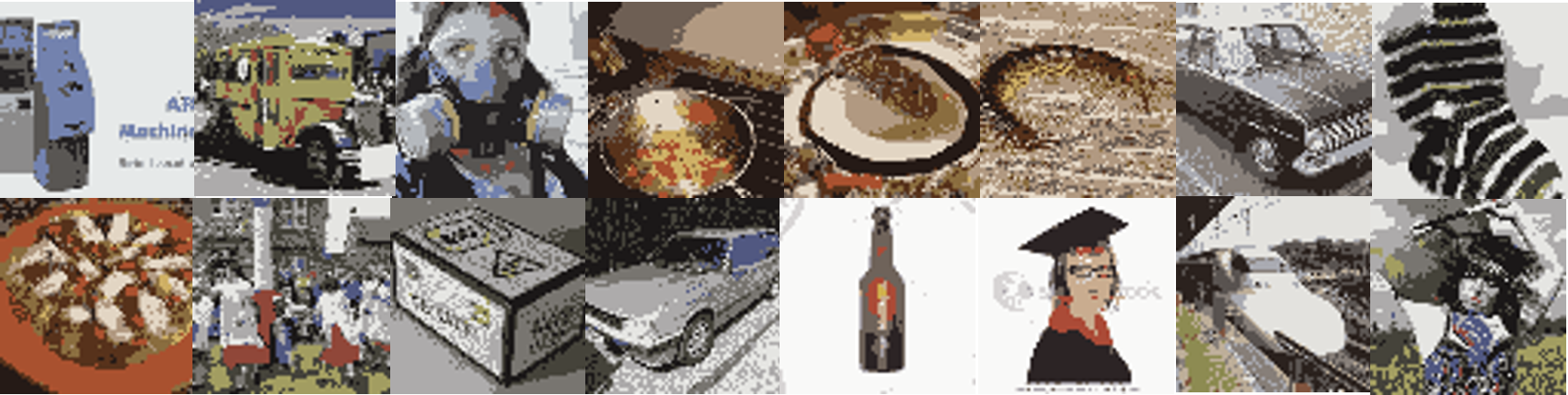}
        \caption{3 Bit (8 colors), Tiny-ImageNet}
        \label{fig:img3}
    \end{subfigure}
    
    \caption{Visualization of Tiny-ImageNet under different color quantization ratios. From top to bottom: 2-bit quantization (4 colors) and 3-bit quantization (8 colors).}
    \label{fig:Tiny-ImageNet-4}
\end{figure}

% \section{Another Visualization}
Figure~\ref{fig:quantization-comparison-cifar} and \ref{fig:Tiny-ImageNet-4} present the visualizations of CIFAR-10/100 and Tiny-ImageNet under different color quantization rates, respectively. 

\end{document}